\documentclass{article}

\usepackage[final,nonatbib]{neurips_2021}

\usepackage[utf8]{inputenc} 
\usepackage[T1]{fontenc}    
\usepackage{hyperref}       
\usepackage{url}            
\usepackage{booktabs}       
\usepackage{amsfonts}       
\usepackage{nicefrac}       
\usepackage{microtype}      
\usepackage{xcolor}         
\usepackage{graphicx}
\usepackage{caption}
\usepackage{subcaption}
\usepackage{amsmath,amsthm,amssymb,amsfonts,bm}
\usepackage{mathtools}
\usepackage{commath}
\usepackage{float}
\usepackage[ruled,vlined,noresetcount,algosection]{algorithm2e}

\SetKwInput{KwRequire}{Require}

\DeclareMathOperator{\tr}{Tr}

\usepackage{hyperref}

\title{AugmentedPCA: A Python Package of \\Supervised and Adversarial Linear Factor Models}

\author{
William E. {Carson IV} \\
Department of Biomedical Engineering \\
Duke University \\
Durham, NC 27708 \\
\texttt{william.carson@duke.edu} \\
\And
Austin {Talbot} \\
Department of Psychiatry and Behavioral Sciences \\
Stanford University \\
Stanford, CA 94305 \\
\texttt{abt23@stanford.edu} \\
\And
David {Carlson} \\
Department of Civil and Environmental Engineering \\
Department of Biostatistics and Bioinformatics \\
Duke University \\
Durham, NC 27708 \\
\texttt{david.carlson@duke.edu} \\
}

\begin{document}

\maketitle

\begin{abstract}

Deep autoencoders are often extended with a supervised or adversarial loss to learn latent representations with desirable properties, such as greater predictivity of labels and outcomes or fairness with respects to a sensitive variable. Despite the ubiquity of supervised and adversarial deep latent factor models, these methods should demonstrate improvement over simpler linear approaches to be preferred in practice. This necessitates a reproducible linear analog that still adheres to an augmenting supervised or adversarial objective. We address this methodological gap by presenting methods that augment the principal component analysis (PCA) objective with either a supervised or an adversarial objective and provide \emph{analytic and reproducible} solutions. We implement these methods in an open-source Python package, AugmentedPCA, that can produce excellent real-world baselines. We demonstrate the utility of these factor models on an open-source, RNA-seq cancer gene expression dataset, showing that augmenting with a supervised objective results in improved downstream classification performance, produces principal components with greater class fidelity, and facilitates identification of genes aligned with the principal axes of data variance with implications to development of specific types of cancer.
\end{abstract}

\section{Introduction}

Dimensionality reduction techniques are widely used in machine learning pipelines, whether for preprocessing purposes or to reveal latent data structure. One common application of dimensionality reduction techniques is in next-gen gene sequencing (NGS) pipelines. NGS methods such as RNA-sequencing (RNA-seq) provide insights into the cellular transcriptome, in turn providing crucial information on connecting gene expression information to phenotype. The expression of this genetic information, in addition to environmental factors, characterizes the phenotype of an organism \cite{kukurba2015rna}. Due to the vast number of genes in the human genome, gene sequencing counts result in extremely high-dimensional data. Dimensionality reduction techniques such as principal component analysis (PCA), t-distributed stochastic neighbor embedding (t-SNE) \cite{van2008visualizing}, or Uniform Manifold Approximation and Projection (UMAP) \cite{mcinnes2018umap} are often used in gene expression analysis pipelines to visualize clustering of samples in 2-dimensional (2D) space or as a preprocessing step for downstream classification. However, sometimes factors and principal components may be representative of global variance aligned with individual differences, rather than variance specific to condition or disease. This could be addressed by implementing factor models with an additional objective that enforces latent components to be predictive of condition or components that are invariant to patient-specific variation. This is commonly accomplished by using deep autoencoder variants trained via stochastic gradient descent (SGD) to optimize a multi-objective loss; however, these methods typically require large amounts of data to generalize well and do not always converge to consistent representations or robust solutions.

Here, we introduce AugmentedPCA, a Python package of linear factor models that provides supervised and adversarial dimensionality reduction. These factor models produce representations aligned with an \textit{augmenting objective} in addition to the canonical PCA objective of finding components that represent the original data variance. AugmentedPCA provides implementations of two factor models: a supervised factor model, which is fit according to a joint objective that enforces greater class fidelity in learned components, and an adversarial factor model, which is fit via an adversarial objective that enforces invariance to concomitant data. Remarkably, these objective functions yield analytic solutions by performing an eigendecomposition over an augmented space, thus providing reproducible and effective linear methods. These techniques are implemented in an open-source Python package to provide easily-implemented baselines.

\section{AugmentedPCA Package Overview}

The AugmentedPCA package provides Python implementations of two separate linear factor models: supervised AugmentedPCA (SAPCA) and adversarial AugmentedPCA (AAPCA). These models have the following objectives:
\begin{itemize}
    \item \textbf{SAPCA}: Find components that i) represent the maximum variance expressed in the primary data (primary objective) while also, ii) representing the variance expressed in the data labels (augmenting objective).
    \item \textbf{AAPCA}: Find components that i) represent the maximum variance expressed in the primary data (primary objective) while, ii) maintaining a degree of invariance to a set of concomitant data (augmenting objective).
\end{itemize}
SAPCA is useful when \textit{predictivity} of latent components with respects to a set of data labels or outcomes is desired. SAPCA is equivalent to a supervised autoencoder (SAE) \cite{ranzato2008semi, zhang2016augmenting} with a single hidden layer. Therefore, SAPCA can be applied to situations where the properties of latent representations enforced via deep SAEs are desired, yet where limited data or training inconsistencies are a concern. AAPCA can be used in situations where one wishes to enforce \textit{invariance} of latent components to a set of concomitant or confounding data, and is equivalent to an adversarial autoencoder \cite{makhzani2015adversarial} with a single hidden layer.

AugmentedPCA model implementations have four key hyperparameters that should be specified when instantiating:
\begin{itemize}
    \item \texttt{n\_components}: Specifies the number of components or latent factors.
    \item \texttt{mu}: Controls the strength of the augmenting objective. A \texttt{mu} value of 0 results in a model equivalent to PCA in which the primary objective of maximizing data variance is prioritized, where higher values increase the emphasis placed on fitting the augmenting objective. In SAPCA, a higher \texttt{mu} value results in factors with greater predictiveness of supervised augmenting data or data labels. In AAPCA, a higher \texttt{mu} value results in factors with greater invariance to concomitant data.
    \item \texttt{inference}: Determines the model inference strategy/how factors are generated. Options include ``local'' and ``encoded''. Further details on inference strategies are given in Section \ref{sec:apca_model_form}.
    \item \texttt{decomp}: Specifies the decomposition approach. Options include ``exact'' and ``approx''. See Section \ref{sec:rapca} and Appendix \ref{sec:rapca_add_details} for more details on decomposition options. 
\end{itemize}

AugmentedPCA models are designed to mimic the scikit-learn PCA implementation in terms of object methods and attributes \cite{buitinck2013api}. The following are key methods for implementing AugmentedPCA models in practice:
\begin{itemize}
    \item \texttt{fit}: This method fits the AugmentedPCA model/calculates model parameters given the primary and augmenting data.
    \item \texttt{transform}: Transforms the data to a latent representation of the specified number of components.
    \item \texttt{reconstruct}: Given the primary and augmenting data, this method provides the reconstructed primary and augmenting data for the specified number of components.
\end{itemize}
The latest stable release of AugmentedPCA can be installed via \texttt{pip} (\texttt{pip install augmented-pca}). AugmentedPCA models do not require GPU acceleration for model fitting or inference. Additionally, the only required dependencies of the AugmentedPCA package are the NumPy and SciPy libraries \cite{harris2020array, virtanen2020fundamental}, thereby facilitating easy integration of AugmentedPCA into most Python virtual environments. AugmentedPCA code can be found at \url{https://github.com/wecarsoniv/augmented-pca} and package documentation can be referenced at \url{https://augmented-pca.readthedocs.io}. AugmentedPCA is available for use under the permissive free software MIT License.

\section{AugmentedPCA Factor Models}
\label{sec:apca_model_form}

Here, we introduce AugmentedPCA model formulations and inference strategies. Derivations of analytic solutions can be referenced in Appendix \ref{sec:derivations}. Further discussion on the properties of these factor models can be found in \cite{talbot2020relating} and \cite{talbot2020supervised}.

\subsection{Review of Principal Components Analysis}

First, we start with a brief of review of PCA. PCA is a linear factor model and dimensionality reduction technique that finds orthogonal components that maximize the explained variance of the data. Let $\mathbf{X} = [\mathbf{x}_1, ..., \mathbf{x}_n] \in \mathbb{R}^{p \times n}$ represent the matrix of $n$ samples or observations of $p$-dimensional de-meaned primary data. The factors or components are denoted by $\mathbf{S} = [\mathbf{s}_1, ..., \mathbf{s}_n] \in \mathbb{R}^{k \times n}$, where $k$ represents the number of chosen factors. The PCA objective function can be expressed as
\begin{equation}
\label{eq:pca_obj}
    \textstyle \min_{\mathbf{W}} \norm{\mathbf{X} - \mathbf{W} \mathbf{S}}_F^2,
\end{equation}
where $\mathbf{W} \in \mathbb{R}^{p \times k}$ represents the loadings matrix and $\norm{\cdot}_F^2$ represents the squared Frobenius norm. The above objective can be solved via an eigendecomposition of the empirical covariance matrix
\begin{equation}
\textstyle 
    \mathbf{B} = \mathbf{X} \mathbf{X}^{\intercal},
\end{equation}
where the solutions for $\mathbf{W}$ are the scaled eigenvectors associated with the $k$ largest eigenvalues. In the following sections, we detail how Equation \ref{eq:pca_obj} can be augmented with a supervised loss to promote greater class fidelity in factors (Section \ref{sup_apca}) or an adversarial loss to create factors that are invariant to concomitant data (Section \ref{sup_apca}).

\subsection{Supervised AugmentedPCA}
\label{sup_apca}

Supervised AugmentedPCA (SAPCA) augments the PCA objective with a supervised loss, similar to an SAE \cite{ranzato2008semi, zhang2016augmenting}. Here, we detail two different approximate inference strategies, termed ``local'' and ``encoded,'' for solving the SAPCA objective.  In this section, $\mathbf{Y} \in \mathbb{R}^{q \times n}$ represents a set of supervision data (e.g., data labels or outcomes) and $\mathbf{D} \in \mathbb{R}^{q \times k}$ represents the linear mapping from the factors to the supervision data.

\subsubsection{Supervised Local Inference}
\label{sec:sup_local_inference}

\begin{figure*}[t!]
  \centering
  \begin{subfigure}[]{0.44\textwidth}
    \centering
    \includegraphics[width=1.0\textwidth]{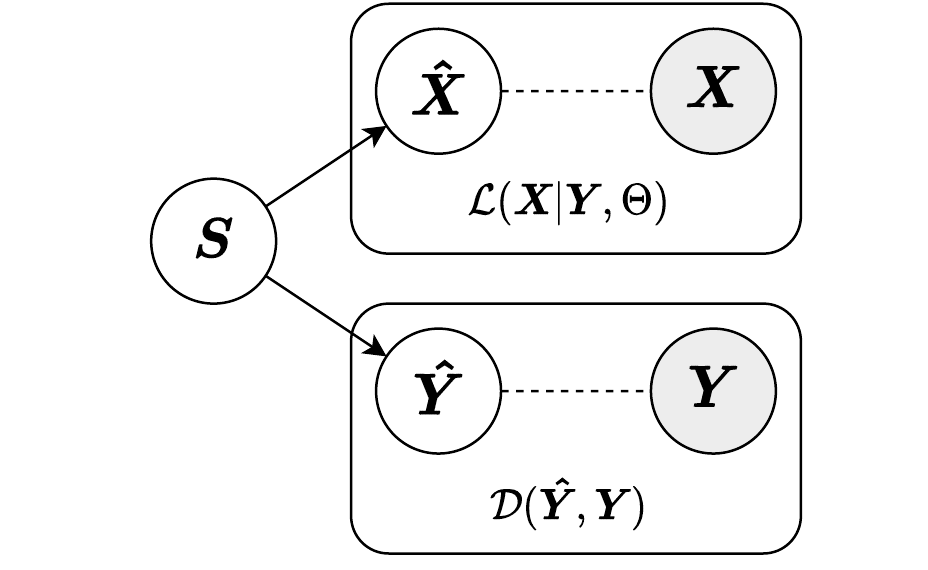}
    \caption{}
    \label{fig:local_inference_diagram}
  \end{subfigure}
  \begin{subfigure}[]{0.52\textwidth}
    \centering
    \includegraphics[width=1.0\textwidth]{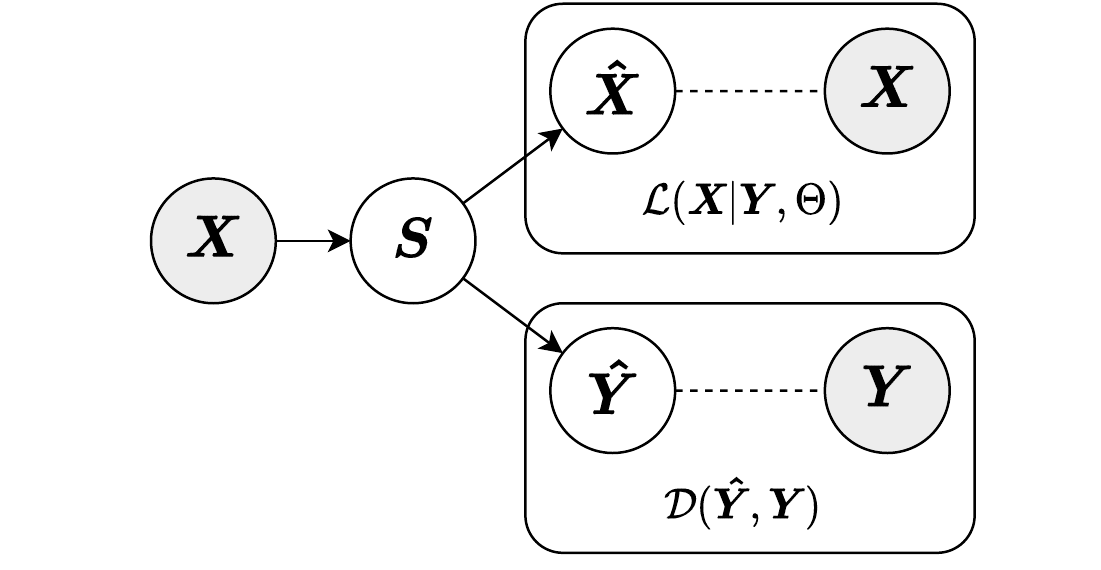}
    \caption{}
    \label{fig:encoded_inference_diagram}
  \end{subfigure}
  \label{fig:inference_strategy_diagram}
  \caption{(a) Diagram depicting the local approximate inference strategy. In the local inference strategy, the factors (local variables associated with each observation) are included explicitly in both the primary and augmenting objective. (b) Diagram depicting the encoded approximate inference strategy. In the encoded inference strategy, a linear encoder (encoding matrix $\mathbf{A}$) is used to transform the data into factors or components. This inference strategy is termed ``encoded'' because the augmenting objective is enforced via an encoding function.}
\end{figure*}

The local inference strategy is based on joint factor models \cite{yu2006supervised} that learn the factors or components from both primary data $\mathbf{X}$ and supervision data $\mathbf{Y}$. The local supervised approximate inference objective function can be expressed as
\begin{equation}
\textstyle
    \min_{\mathbf{W}, \mathbf{D}, \mathbf{S}} \norm{\mathbf{X} - \mathbf{W} \mathbf{S}}_F^2 + \mu \norm{\mathbf{Y} - \mathbf{D} \mathbf{S}}_F^2,
\label{eq:lsapca_loc_obj}
\end{equation}
where $\mu$ represents the strength of the augmenting supervised objective. Since components are learned using information from both $\mathbf{X}$ and $\mathbf{Y}$, local inference should only be used when the augmenting data are known or accessible at test-time. For most classification and prediction problems, we believe that the encoded approach (described in Section \ref{sec:sup_encoded_inference}) is more appropriate, but several use cases of the local approach exist in the joint modeling literature \cite{yu2006supervised}. The solutions for $[\mathbf{W}^{\intercal}, \mathbf{D}^{\intercal}]^{\intercal}$ correspond to the scaled eigenvectors associated with the $k$ largest eigenvalues of the matrix
\begin{equation}
\textstyle
    \mathbf{B}_{\text{SL}} = \textstyle\begin{bmatrix} \mathbf{X} \mathbf{X}^{\intercal} & \mu \mathbf{X} \mathbf{Y}^{\intercal} \\ \mathbf{Y} \mathbf{X}^{\intercal} & \mu \mathbf{Y} \mathbf{Y}^{\intercal} \end{bmatrix}.
\label{eq:sapca_loc_dec_mat}
\end{equation}
A diagram depicting the local inference strategy can be seen in Figure \ref{fig:local_inference_diagram}.

\subsubsection{Supervised Encoded Inference}
\label{sec:sup_encoded_inference}

In the encoded inference strategy, components are estimated exclusively from the information in $\mathbf{X}$ via a linear mapping, the parameters of which are given by the matrix $\mathbf{A} \in \mathbb{R}^{p \times k}$. The encoded supervised approximate inference objective function can be expressed as
\begin{equation}
\textstyle
    \min_{\mathbf{W}, \mathbf{D}, \mathbf{A}} \norm{\mathbf{X} - \mathbf{W} \mathbf{A} \mathbf{X}}_F^2 + \mu \norm{\mathbf{Y} - \mathbf{D} \mathbf{A} \mathbf{X}}_F^2.
\label{eq:sapca_enc_obj}
\end{equation}
Since factors are learned using only the variance explainable by $\mathbf{X}$, encoded inference is preferable when augmenting data are not known at test-time. The solutions for $[\mathbf{W}^{\intercal}, \mathbf{D}^{\intercal}]^{\intercal}$ correspond to the scaled eigenvectors associated with the $k$ largest eigenvalues of the matrix
\begin{equation}
    \mathbf{B}_{\text{SE}} = \begin{bmatrix} \mathbf{X} \mathbf{X}^{\intercal} & \mu \mathbf{X} \mathbf{Y}^{\intercal} \\ \mathbf{Y} \mathbf{X}^{\intercal} & \mu \mathbf{Y} \mathbf{P}_{\mathbf{X}} \mathbf{Y}^{\intercal} \end{bmatrix},
\label{eq:sapca_enc_dec_mat}
\end{equation}
where $\mathbf{P}_{\mathbf{X}} = \mathbf{X}^{\intercal} (\mathbf{X} \mathbf{X}^{\intercal})^{-1} \mathbf{X}^{\intercal}$ is the projection onto $\mathbf{X}$. Once solutions to $\mathbf{W}$ and $\mathbf{D}$ are obtained they can be used to calculate $\mathbf{A}$ via a closed-form linear algebra equation, which is detailed in Appendix \ref{sec:derivations}. A diagram depicting the encoded inference strategy can be seen in Figure \ref{fig:encoded_inference_diagram}.

\subsection{Adversarial AugmentedPCA}
\label{adv_apca}

Adversarial AugmentedPCA (AAPCA) is preferable when we wish to learn a factor representation $\mathbf{S}$ that is not overly-predictive of concomitant data $\mathbf{Y}$, which is a common goal in the domain adaptation \cite{ganin2016domain} and fairness literature \cite{louizos2016variational}. In this section, we let $\mathbf{Y}$ represent concomitant data and $\mathbf{D}$ represent the linear mapping between the components to the concomitant data.

\subsubsection{Adversarial Local Inference}
\label{sec:adv_local_inference}

The adversarial local approximate inference objective can be expressed as
\begin{equation}
\textstyle
    \min_{\mathbf{W}, \mathbf{D}, \mathbf{S}} \norm{\mathbf{X} - \mathbf{W} \mathbf{S}}_F^2 - \mu \norm{\mathbf{Y} - \mathbf{D} \mathbf{S}}_F^2,
\label{eq:aapca_loc_obj}
\end{equation}
where $\mu$ represents the strength of the augmenting adversarial objective, and where the solutions for $[\mathbf{W}^{\intercal},\mathbf{D}^{\intercal}]^{\intercal}$ correspond to the scaled eigenvectors associated with the $k$ largest eigenvalues of the matrix
\begin{equation}
    \mathbf{B}_{\text{AL}} = \begin{bmatrix} \mathbf{X} \mathbf{X}^{\intercal} & -\mu \mathbf{X} \mathbf{Y}^{\intercal} \\ \mathbf{Y} \mathbf{X}^{\intercal} & -\mu \mathbf{Y} \mathbf{Y}^{\intercal} \end{bmatrix}.
\label{eq:aapca_loc_dec_mat}
\end{equation}
This ``local'' approach requires that we know/have access to concomitant information when projecting to the latent space (e.g., at test time), and can be viewed as correcting the latent space for the concomitant information. This is the approach taken by much of the fair machine learning literature \cite{louizos2016variational}.

\subsubsection{Adversarial Encoded Inference}
\label{sec:adv_encoded_inference}

The encoded adversarial inference strategy objective can be expressed as
\begin{equation}
\textstyle
    \min_{\mathbf{W}, \mathbf{D}, \mathbf{A}} \norm{\mathbf{X} - \mathbf{W} \mathbf{A} \mathbf{X}}_F^2 - \mu \norm{\mathbf{Y} - \mathbf{D} \mathbf{A} \mathbf{X}}_F^2,
\label{eq:aapca_enc_obj}
\end{equation}
where $\mathbf{A}$ represents the linear encoding matrix as in Section \ref{sec:sup_encoded_inference}.
The solutions for $[\mathbf{W}^{\intercal},\mathbf{D}^{\intercal}]^{\intercal}$ correspond to the scaled eigenvectors associated with the $k$ largest eigenvalues of the matrix
\begin{equation}
    \mathbf{B}_{\text{AE}} = \begin{bmatrix} \mathbf{X} \mathbf{X}^{\intercal} & - \mu \mathbf{X} \mathbf{Y}^{\intercal} \\ \mathbf{Y} \mathbf{X}^{\intercal} & - \mu \mathbf{Y} \mathbf{P}_{\mathbf{X}} \mathbf{Y}^{\intercal} \end{bmatrix}.
\label{eq:aapca_enc_dec_mat}
\end{equation}
The matrix $\mathbf{A}$ can be calculated via a closed-form linear algebra equation using $\mathbf{W}$ and $\mathbf{D}$, which is detailed in Appendix \ref{sec:derivations}. This ``encoded'' approach does not require that the concomitant information is known when projecting to the latent space, and is appropriate when we would not expect to know the information (e.g., confidential or protected information).

\subsection{Randomized AugmentedPCA for Tractable Inference}
\label{sec:rapca}

One limitation of the analytic AugmentedPCA formulations is that they require eigendecompositions which have a complexity of $\mathcal{O}(P^3)$, where $P$ represents the total number of features (combined number of primary and augmenting features). The challenge of polynomial complexity can be addressed by leveraging random projections to quickly compute approximate matrix decompositions \cite{halko2011finding}. Algorithm \ref{alg:random_eig} describes how to use a random matrix in a subspace iteration algorithm for approximating the eigenvalues of a square matrix. This approximate fit technique greatly decreases algorithmic complexity from $\mathcal{O}(P^3)$ in the exact eigendecomposition case to $\mathcal{O}(P^2 (k + s))$ for the randomized approximation, where $k$ represents the number of factors or components and $s$ represents an oversampling parameter (see Appendix \ref{sec:rapca_add_details} for further details on the randomized algorithm used for computing approximate eigendecompositions). This algorithm is incorporated as a model fit option in AugmentedPCA implementations and can be chosen by passing ``approx'' as the argument to the $\texttt{decomp}$ model parameter during AugmentedPCA model instantiation. In practice, using an approximate decomposition for AugmentedPCA models may be preferred for extremely high-dimensional data (e.g., $>$ 10,000 features).

\begin{figure*}[t!]
  \centering
  \includegraphics[width=0.69\textwidth]{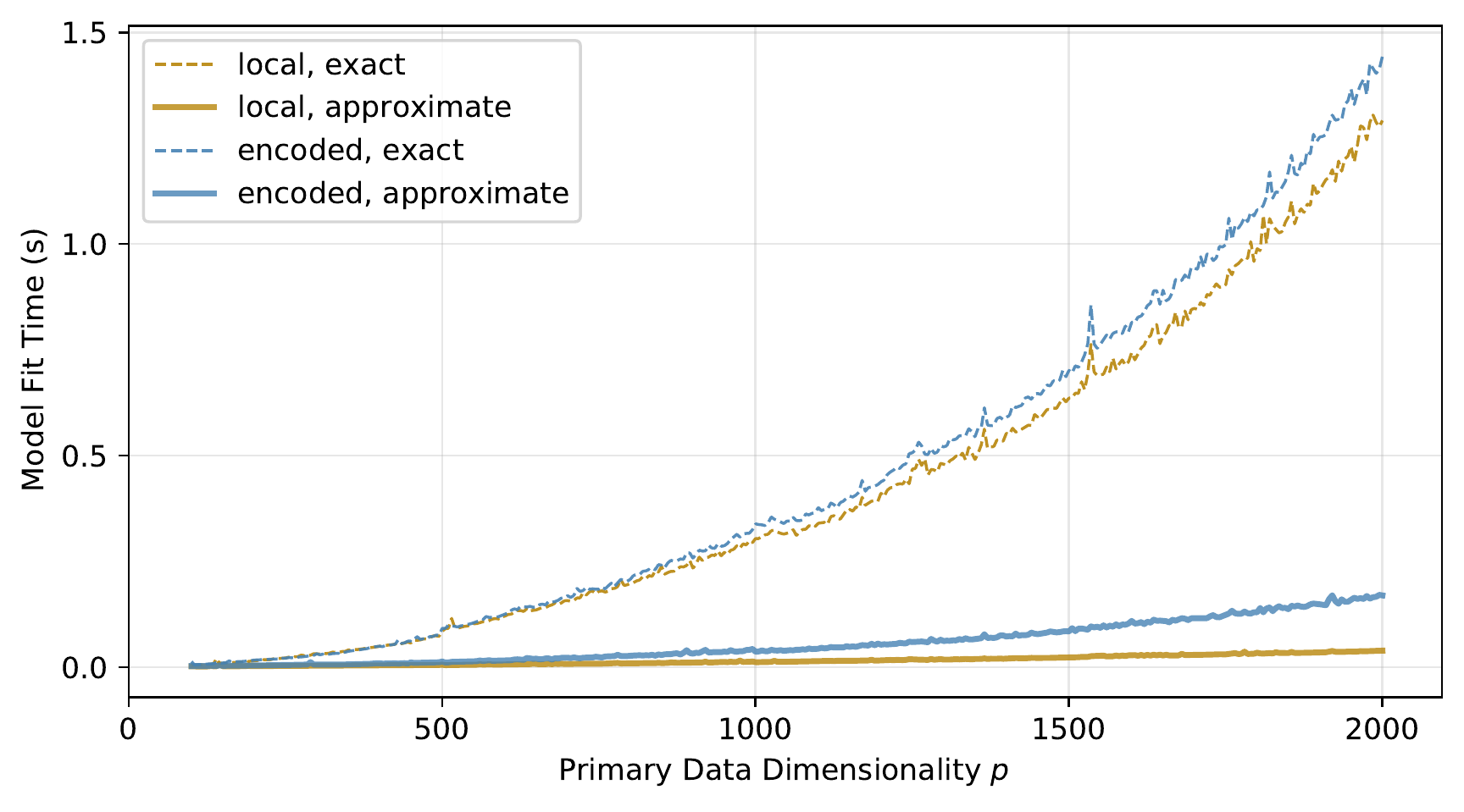}
  \caption{Wall-clock time of exact (solid lines) and approximate (dashed lines) model fit strategies as a function of increasing primary data dimensionality $p$. Number of data samples and augmenting data dimensionality were kept constant at $n$ = 1,000 and $q$ = 1, respectively. For approximate model fits, 5 power iterations and oversampling value of 5 were used. Both primary and augmenting data were randomly generated from a standard normal distribution for each value of $p$.}
  \label{fig:model_fit_time}
\end{figure*}

This randomized approach produces factors that well-approximate the analytic AugmentedPCA model solutions while considerably reducing model fit time. Figure \ref{fig:model_fit_time} displays wall-clock time required to fit AugmentedPCA models as a function of increasing number of primary data features $p$ for both exact and approximate model fit strategies. Exact model fit time plot trajectories (depicted as dashed lines) are indicative of polynomial time complexity, which aligns with the algorithmic complexity of the eigendecomposition bottleneck. Conversely, while approximate model fit time plot trajectories (depicted as thick, solid lines) also increase as a function of increasing number of features, the fit times are significantly less than that of their corresponding exact fit analog.

\section{Supervised AugmentedPCA Example - Cancer Gene Expression Data}
\label{sec:sup_apca_example}

In this section, we demonstrate the utility of AugmentedPCA models by using SAPCA on a \href{https://archive.ics.uci.edu/ml/datasets/gene+expression+cancer+RNA-Seq}{open-source cancer gene expression dataset} from the UCI machine learning repository \cite{weinstein2013cancer}. This dataset contains deidentified RNA-seq gene expression levels measured from patients diagnosed with one of five different types of cancers: breast cancer (BRCA), colon adenocarcinoma (COAD), kidney renal clear cell carcinoma (KIRC), lung adenocarcinoma (LUAD), and prostate adenocarcinoma (PRAD). For all experiments, classification was performed using a logistic regression classifier and AugmentedPCA model parameters were fit using a computer with an Intel i7-10700K processor. No GPU acceleration was necessary for AugmentedPCA model fitting. An additional visual use case of AAPCA on natural image data is provided in Appendix \ref{sec:adv_apca_example}. 

\subsection{Component Downstream Classification}
\label{sec:downstream_classification}

First, we demonstrate the ability of SAPCA to generate latent components with greater downstream classification performance and generalization to typical linear factor models such as PCA. RNA-seq counts for 20,531 genes were provided for 801 tumor samples, resulting in a primary data matrix $\mathbf{X}^{\intercal} \in \mathbb{R}^{801 \times 20,531}$. Primary data features were scaled to have zero mean and unit variance based on the training data. Cancer type labels were transformed into one-hot encodings for each sample, resulting in a supervision data matrix $\mathbf{Y}^{\intercal} \in \mathbb{R}^{801 \times 5}$. The data was subsequently divided into a roughly 50-50 train-test split using the \texttt{train\_test\_split()} function from the scikit-learn package \cite{pedregosa2011scikit}. We use SAPCA with 2 components to create representations with greater fidelity with respects to cancer type. Encoded inference is used as the approximate inference strategy since access to labels at test time is not appropriate given the components use in a downstream classification task. A logistic regression classifier was fit using the components derived from the training data and tested on the components derived from the test data for supervision strengths $\mu$ = 0, 100, ..., 40,000.

\begin{figure*}[t!]
  \begin{subfigure}[]{0.44\textwidth}
    \begin{subfigure}[]{0.99\textwidth}
        \centering
        \includegraphics[width=0.95\textwidth]{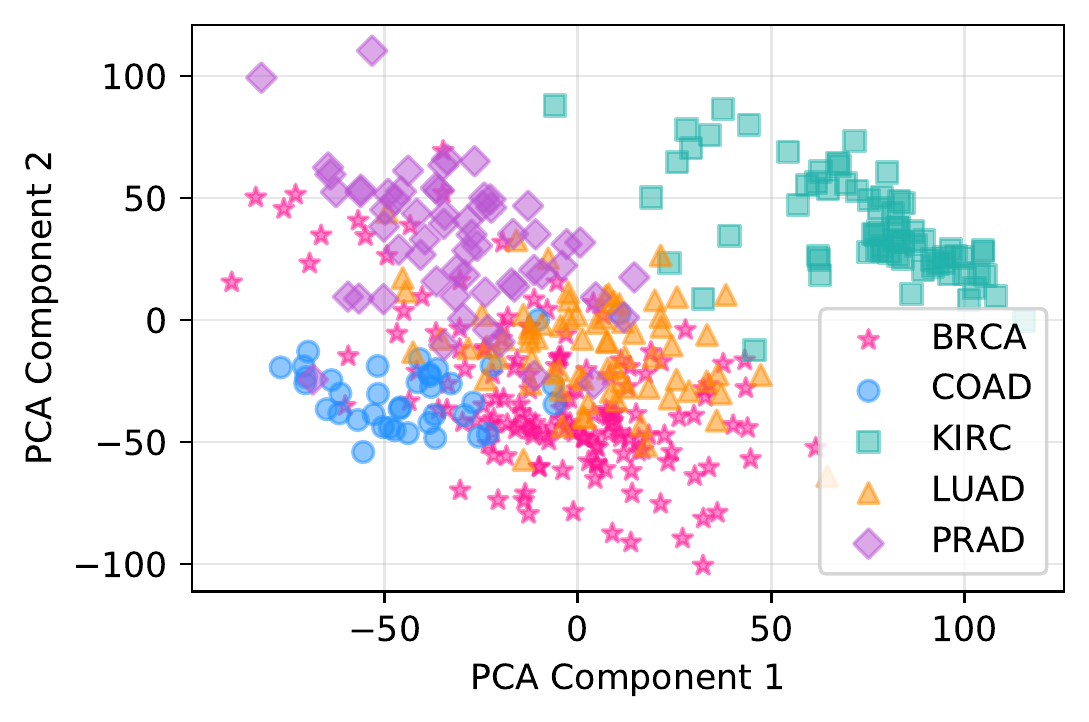}
        \caption{}
        \label{fig:gene_pca_cluster}
    \end{subfigure}
    \begin{subfigure}[]{0.99\textwidth}
        \centering
        \includegraphics[width=0.95\textwidth]{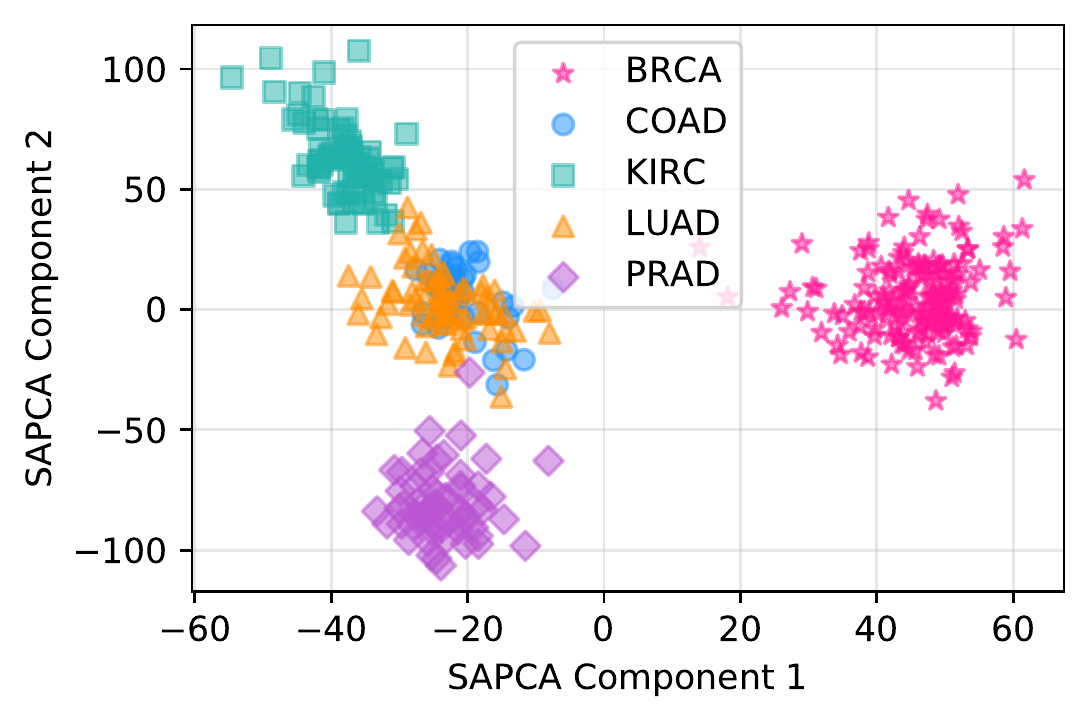}
        \caption{}
        \label{fig:gene_SAPCA_cluster}
    \end{subfigure}
  \end{subfigure}
  \begin{subfigure}[]{0.54\textwidth}
    \centering
    \includegraphics[width=0.95\textwidth]{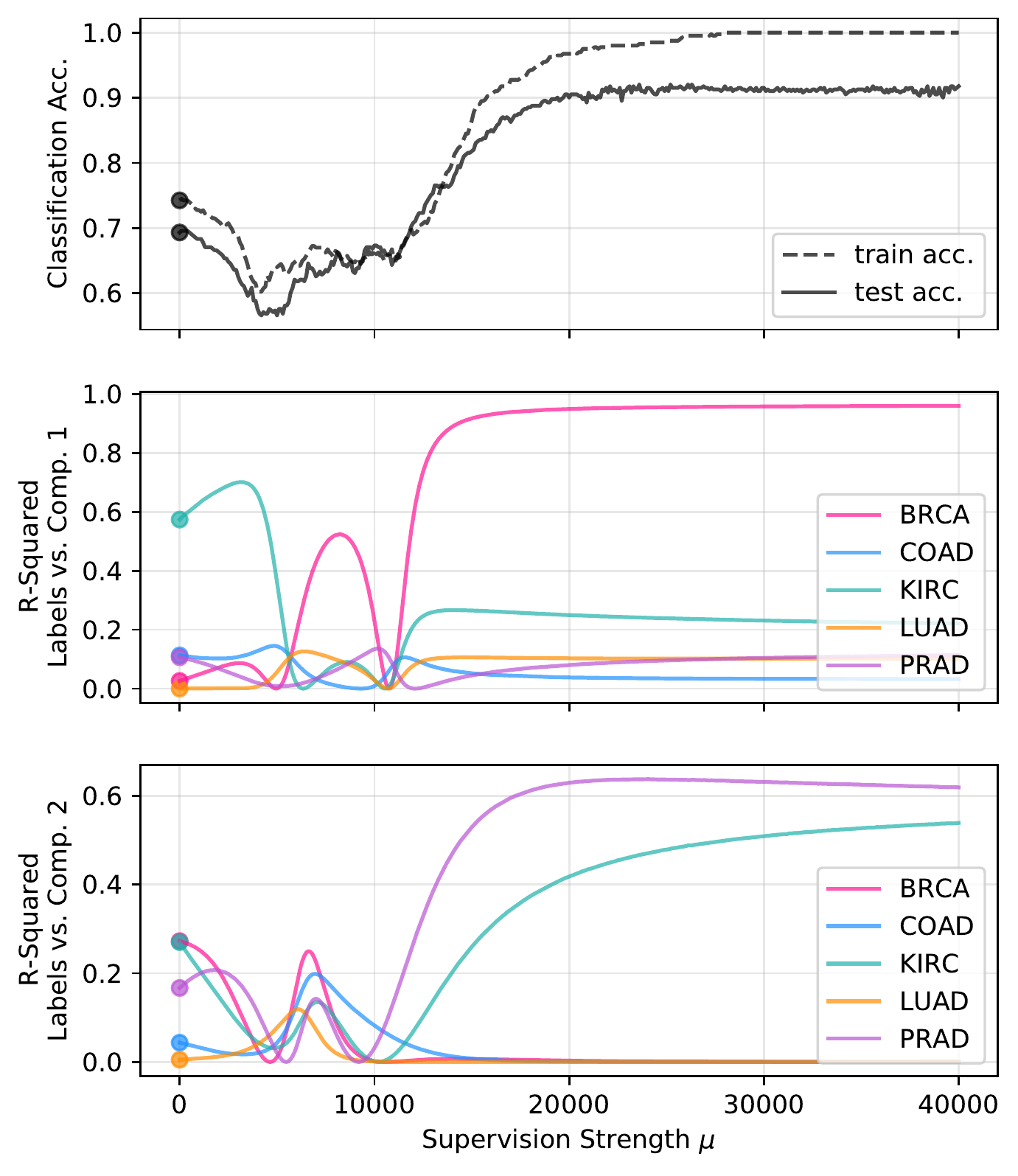}
    \caption{}
    \label{fig:cancer_classification}
  \end{subfigure}
  \caption{(a) PCA clustering of test set cancer gene expression samples. (b) SAPCA clustering of test set cancer gene expression samples with supervision strength $\mu$ = $4 \cdot 10^{4}$. (c) Top: tumor classification accuracy as a function of supervision strength, middle: R-squared between test set one-hot encoded class labels and test set component 1 scores, bottom: R-squared between test set one-hot encoded class labels and test set component 2 scores. Plotted dots represent classification accuracy and R-squared corresponding to initial PCA solution when $\mu$ = 0.}
\end{figure*}

We visualize 2D clustering of cancer gene expression samples resulting from PCA decomposition in Figure \ref{fig:gene_pca_cluster} and SAPCA decomposition in Figure \ref{fig:gene_SAPCA_cluster}. SAPCA components provide much greater class separation in 2D space compared to the standard PCA approach. These SAPCA components with greater class alignment also lend to better downstream classification of tumor type, as seen in the top-most plot of Figure \ref{fig:cancer_classification}. Test set tumor classification is 71\% when using just two PCA components. At lower supervision strengths, the classification accuracy decreases as components are realigned to emphasize the supervised objective, despite the fact that the class loss is decreasing. However, at higher supervision strengths the test set tumor classification accuracy reaches 92\% when using two SAPCA components as predictors.

To provide insight into the transformation of SAPCA components as a function of supervision strength, we visualize the R-squared between SAPCA component scores and cancer labels in the middle and bottom-most plot of Figure \ref{fig:cancer_classification}. As supervision strength increases, the R-squared between component 1 scores and BRCA labels saturates near 1. This is reflected in Figure \ref{fig:gene_SAPCA_cluster}, as BRCA test set samples achieve clear linear separation between all other cancer test set samples in 2D SAPCA space. Similarly, increasing R-squared between SAPCA component 2 scores and PRAD and KIRC labels with increasing supervision strength is depicted in the bottom plot of Figure \ref{fig:cancer_classification}. The alignment of the labels of these two types of cancer with the second component manifests in improved separation of PRAD and KIRC test samples from all other cancers along the second component axis in Figure \ref{fig:gene_SAPCA_cluster}. 

\subsection{AugmentedPCA Loadings Interpretation}
\label{sec:loadings_interpretation}

In addition to downstream classification, learned AugmentedPCA representations can be interpreted due to the linearity of the models. In this section, we learned SAPCA representations specific to each of the five different types of cancer in the RNA-seq dataset by enforcing components to be predictive of binary cancer labels for each of the five cancer types. Here, the augmenting data corresponds to a matrix $\mathbf{Y}^{\intercal} \in \mathbb{R}^{801 \times 1}$ for a given cancer type. After fitting SAPCA models, elements of the loading vector corresponding to the first principal component were sorted according to descending magnitudes to identify the top genes that were implicated in each specific type of cancer.

\begin{figure*}[t!]
  \centering
  \begin{subfigure}[]{0.486\textwidth}
    \centering
    \includegraphics[width=0.98\textwidth]{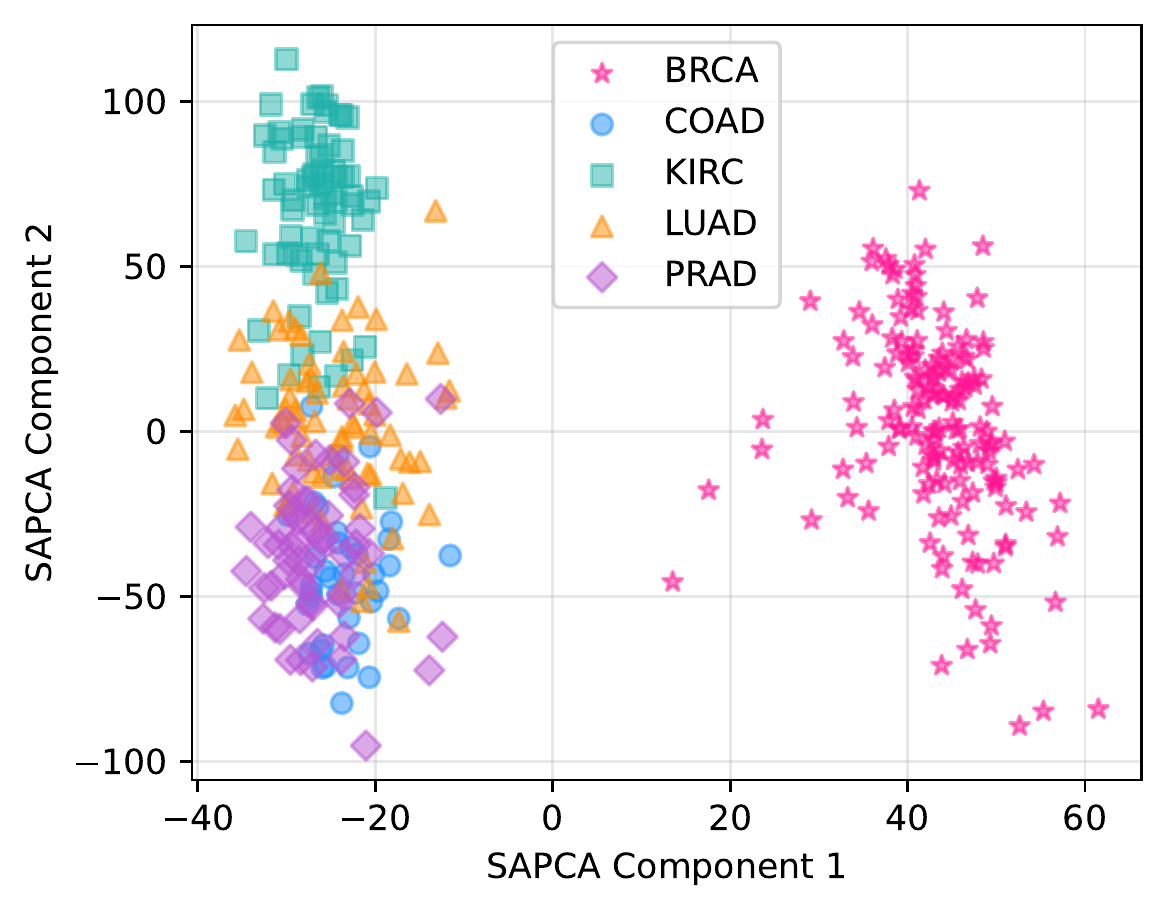}
    \caption{}
    \label{fig:SAPCA_cluster_brca}
  \end{subfigure}
  \begin{subfigure}[]{0.474\textwidth}
    \centering
    \includegraphics[width=0.98\textwidth]{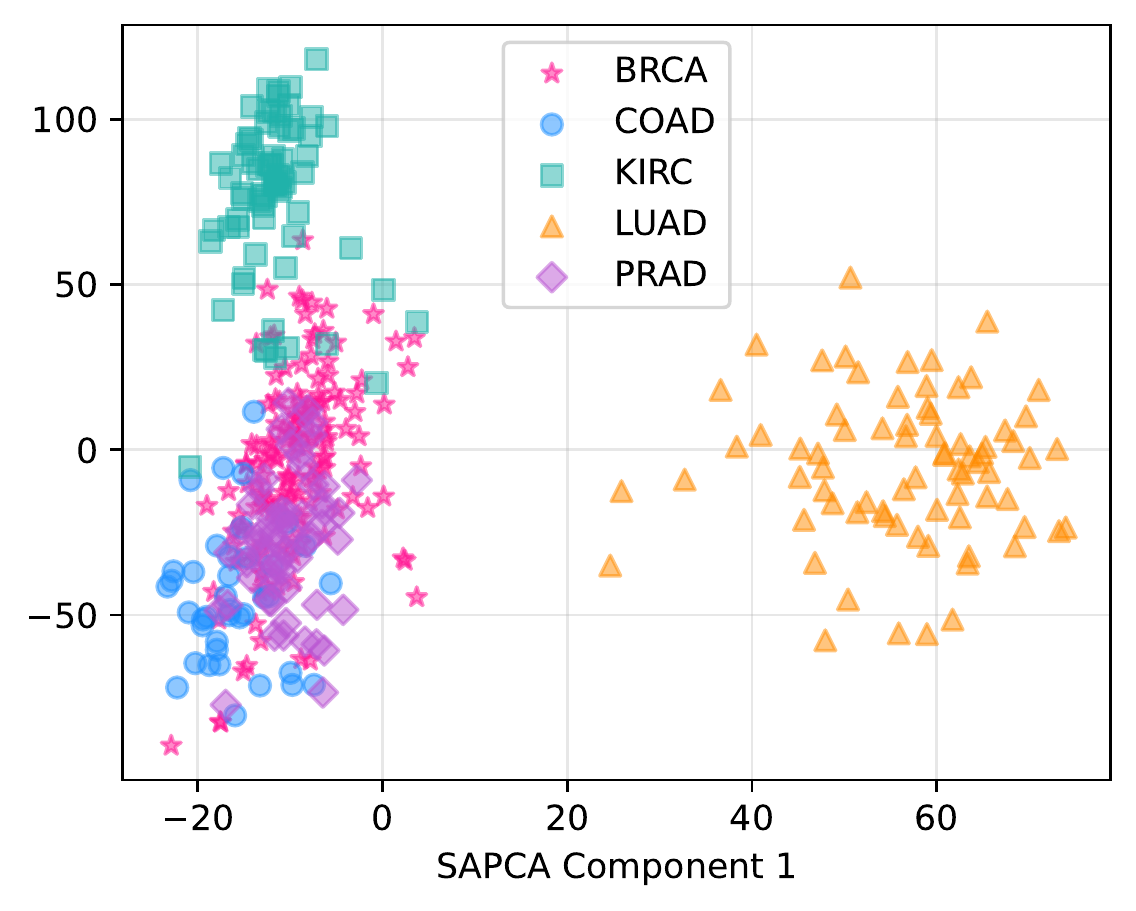}
    \caption{}
    \label{fig:SAPCA_cluster_luad}
  \end{subfigure}
  \caption{(a) Clustering of test set samples with SAPCA components enforced to be predictive of BRCA labels. Supervision strength $\mu$ = $4 \cdot 10^{4}$. (b) Clustering of test set samples with SAPCA components enforced to be predictive of LUAD labels. Supervision strength $\mu$ = $4 \cdot 10^{4}$.}
\end{figure*}

In Table 1, we list the genes corresponding to the highest element magnitudes of SAPCA component 1 loading when enforcing components to be predictive of BRCA only. Literature searches of the genes listed in Table 1 reveal that most are implicated in the development of breast cancers. For instance, TRPS1 is highly-expressed in breast cancers \cite{chang2004trps1, cornelissen2020trps1, elster2018trps1, hu2014trps1, radvanyi2005gene, witwicki2018trps1, wu2014central} and has been used as a specific biomarker for identifying triple-negative breast carcinomas \cite{ai2021trps1}. Additionally, the sign of loading values can be interpreted with respects to the corresponding gene and gene function as well. For example, the FOXF1 gene exhibits tumor-suppressive properties, and epigenetic \textit{inactivation} of FOXF1 has been shown to be associated with breast cancer development. Thus, the negative sign on the loading value aligns with the finding silencing of FOXF1 expression is associated with breast cancers. Interestingly, the top gene implicated in BRCA according to learned SAPCA representations is HNF1B, on which there is not much literature supporting specific implication in development of breast cancers; however, HNF1B has been identified as a biomarker for ovarian cancer risk \cite{shen2013epigenetic}. Figure \ref{fig:SAPCA_cluster_brca} shows test set samples clustered in 2D SAPCA space. We note that there is clear separation between BRCA samples and all other cancers, demonstrating that SAPCA learns a robust principal component basis that separates BRCA samples from all other cancers.

\begin{table}
    \centering
    \caption{Top 10 genes associated with BRCA according to learned SAPCA representations. SAPCA was used to learn factors that were predictive of BRCA labels while also explaining the variance of the full RNA-seq gene expression data. Genes were sorted according to the largest magnitude elements of SAPCA component 1 loading.}
    \smallskip
    \begin{tabular}{|c|c|c|}
    \hline
    \textbf{Gene Symbol} & \textbf{Location}  & \textbf{Loading Value} \\
    \hline
    HNF1B & chr17: 37.68-37.74 Mb & -0.0258 \\
    TRPS1 & chr8: 115.41-115.81 Mb & 0.0250 \\
    TCF21 & chr6: 133.89-133.90 Mb & -0.0250 \\
    GATA3 & chr10: 8.05-8.08 Mb & 0.0246 \\
    LMX1B & chr9: 126.61-126.70 Mb & 0.0242 \\
    FOXF1 & chr16: 86.51-86.52 Mb & -0.0239 \\
    FZD5 & chr2: 207.75-207.77 Mb & -0.0239 \\
    SCGB2A2 & chr11: 62.270-62.273 Mb & 0.0238 \\
    PRLR & chr5: 35.05-35.23 Mb & 0.0232 \\
    GATA3-AS1 & chr10: 8.02-8.05 Mb & 0.0231 \\
    \hline
    \end{tabular}
    \label{tab:top_genes_brca}
\end{table}

\begin{table}
    \centering
    \caption{Top 10 genes associated with LUAD according to learned SAPCA representations. SAPCA was used to learn factors that were predictive of LUAD labels while also explaining the variance of the full RNA-seq gene expression data. Genes were sorted according to the largest magnitude elements of SAPCA component 1 loading.}
    \smallskip
    \begin{tabular}{|c|c|c|}
    \hline
    \textbf{Gene Symbol} & \textbf{Location}  & \textbf{Loading Value}\\
    \hline
    SFTPB & chr2: 85.66-85.67 Mb & 0.0338 \\
    SFTA3 & chr14: 36.47-36.52 Mb & 0.0331 \\
    SFTA2 & chr6: 30.8991-30.89995 Mb & 0.0328 \\
    SFTPA1 & chr10: 79.61-79.62 Mb & 0.0325 \\
    SCGB3A2 & chr5: 147.87-147.88 Mb & 0.0324 \\
    ROS1 & chr6: 117.29-117.43 Mb & 0.0323 \\
    SFTPC & chr8: 22.156-22.164 Mb & 0.0314 \\
    MUC21 & chr6: 30.98-30.99 Mb & 0.0313 \\
    NKX2-1 & chr14: 36.516-36.521 Mb & 0.0310 \\
    SCEL & chr13: 77.54-77.65 Mb & 0.0307 \\
    \hline
    \end{tabular}
    \label{tab:top_genes_luad}
\end{table}

Similarly, in Table 2 we list the genes corresponding to the highest SAPCA component 1 loading magnitudes when enforcing components to be predictive of LUAD only. Notably, four of the top ten genes learned by SAPCA to be predictive of lung cancer are all surfactant-associated genes (``SF'' prefix), with many of these genes having been specifically identified as biomarkers for lung cancer risk \cite{sin2013pro, xiao2017eight, wang2009genetic} or associated with development of pulmonary diseases such as interstitial pneumonias \cite{honda2018deleterious, nathan2016germline}. Again in Figure \ref{fig:SAPCA_cluster_luad}, we note a clear separation between LUAD samples and all other cancers when visualizing samples in 2D SAPCA space.

We repeated this process for all cancer types in the dataset. Tables of top 10 genes implicated in COAD, KIRC, and PRAD according to learned SAPCA representations can be seen in Appendix \ref{sec:loadings_interpretation_additional}.

\section{Discussion and Conclusion}
\label{sec:discussion_conclusion}

Linear methods are often favored as baseline approaches over more complex deep learning methodology due to their reproducibility, robustness, and inherent interpretability. While the factor model approach is not strictly necessary for prediction, it maintains a linear model that provides interpretable factor loadings and data reconstructions. Additionally, recent work has demonstrated that including reconstruction error with a supervised classification objective results in improved generalization \cite{le2018supervised}. Thus, the inclusion of reconstruction error in AugmentedPCA model objectives may provide better generalization in downstream classification tasks.



Additionally, linear factor models such as supervised AugmentedPCA models provide a benefit over linear classification methods such as logistic regression in identifying features important features if the features express multicollinearity. Linear classification methods such as logistic regression assume that there is no multicollinearity between features, i.e., that features are more or less independent from one another. Multicollinear features violate the assumptions of linear classification models and may result in poor generalization. Gene expression levels from the cancer gene expression dataset used in this work are an example of features that express multicollinearity as genes from the same gene network or adjacent loci may be subject to similar epigenetic modification profiles or have similar chromatin accessibility, thereby resulting in multicollinear expression. Thus, linear factor models such as AugmentedPCA provide a benefit over logistic regression in identifying multiple genes from adjacent loci, potentially aiding in identification of possible sites of epigenetic modification for expression and gene pathways for future exploration. For example, SAPCA identifies two genes, SFTA2 and MUC21, from the adjacent loci on chromosome 6 as two of the top 10 most important genes in development of LUAD.

One limitation of the AugmentedPCA models is that they are linear in nature, whereas deep learning approaches have shown improved performance in many complex big data tasks. However, these complex techniques should show real improvements over simpler, reproducible baselines to be preferred in practice, motivating the adoption of AugmentedPCA models as baseline methods. Additionally, while deep learning methods may work for larger datasets, they may fail to learn representations that are robust to domain shift for smaller datasets with only a few hundred points. In these cases of small data, a linear model with analytic decompositions, such as the AugmentedPCA models described here, can naturally apply with stable, reproducible results.

In summary, our proposed package of linear factor models, AugmentedPCA, provides a simple and straightforward method against which to benchmark results for new methods of supervised and adversarial representation learning, as well as linear dimensionality reduction techniques that facilitate high-dimensional data interpretation and downstream classification.

\section*{Acknowledgments}
\small
Research reported in this manuscript was supported by the National Institute of Biomedical Imaging and Bioengineering and the National Institute of Mental Health through the National Institutes of Health BRAIN Initiative under Award Number R01EB026937. The contents of this manuscript are solely the responsibility of the authors and do not necessarily represent the official views of any of the funding agencies or sponsors.

{\small
\bibliographystyle{plain}
\bibliography{main.bib}}

\begin{thebibliography}{10}

\bibitem{ai2021trps1}
Di~Ai, Jun Yao, Fei Yang, Lei Huo, Hui Chen, Wei Lu, Luisa Maren~Solis Soto,
  Mei Jiang, Maria~Gabriela Raso, Shufang Wang, et~al.
\newblock {TRPS1}: {A} highly sensitive and specific marker for breast
  carcinoma, especially for triple-negative breast cancer.
\newblock {\em Modern Pathology}, 34(4):710--719, 2021.

\bibitem{buitinck2013api}
Lars Buitinck, Gilles Louppe, Mathieu Blondel, Fabian Pedregosa, Andreas
  Mueller, Olivier Grisel, Vlad Niculae, Peter Prettenhofer, Alexandre
  Gramfort, Jaques Grobler, Robert Layton, Jake VanderPlas, Arnaud Joly, Brian
  Holt, and Ga{\"{e}}l Varoquaux.
\newblock {API} {D}esign for {M}achine {L}earning {S}oftware: {E}xperiences
  from the scikit-learn {P}roject.
\newblock In {\em European Conference on Machine Learning and Principles and
  Practices of Knowledge Discovery in Databases}, pages 108--122, 2013.

\bibitem{chang2004trps1}
Glenn~TG Chang, Mila Jhamai, Wytske~M van Weerden, Guido Jenster, and Albert~O
  Brinkmann.
\newblock {T}he {TRPS1} transcription factor: androgenic regulation in prostate
  cancer and high expression in breast cancer.
\newblock {\em Endocrine-Related Cancer}, 11(4):815--822, 2004.

\bibitem{cornelissen2020trps1}
Lisette~M. Cornelissen, Anne~Paulien Drenth, Eline Van Der~Burg, Roebi
  De~Bruijn, Colin~E.J. Pritchard, Ivo~J. Huijbers, Wilbert Zwart, and Jos
  Jonkers.
\newblock {TRPS1} acts as a context-dependent regulator of mammary epithelial
  cell growth/differentiation and breast cancer development.
\newblock {\em Genes \& Development}, 34(3-4):179--193, 2020.

\bibitem{elster2018trps1}
Dana Elster, Marie Tollot, Karin Schlegelmilch, Alessandro Ori, Andreas
  Rosenwald, Erik Sahai, and Bj{\"o}rn von Eyss.
\newblock {TRPS1} shapes {YAP}/{TEAD}-dependent transcription in breast cancer
  cells.
\newblock {\em Nature Communications}, 9(1):1--16, 2018.

\bibitem{ganin2016domain}
Yaroslav Ganin, Evgeniya Ustinova, Hana Ajakan, Pascal Germain, Hugo
  Larochelle, Fran{\c{c}}ois Laviolette, Mario Marchand, and Victor Lempitsky.
\newblock {D}omain-{A}dversarial {T}raining of {N}eural {N}etworks.
\newblock {\em {T}he {J}ournal of {M}achine {L}earning {R}esearch},
  17(1):2096--2030, 2016.

\bibitem{georghiades2001from}
Athinodoros~S. Georghiades and Peter~N. Belhumeur.
\newblock {F}rom {F}ew to {M}any: {I}llumination {C}one {M}odels for {F}ace
  {R}ecognition under {V}ariable {L}ighting and {P}ose.
\newblock {\em IEEE Transactions on Pattern Analysis and Machine Intelligence},
  23(6):643--660, 2001.

\bibitem{halko2011finding}
Nathan Halko, Per-Gunnar Martinsson, and Joel~A. Tropp.
\newblock {F}inding {S}tructure with {R}andomness: {P}robabilistic {A}lgorithms
  for {C}onstructing {A}pproximate {M}atrix {D}ecompositions.
\newblock {\em SIAM Review}, 53(2):217--288, 2011.

\bibitem{harris2020array}
Charles~R. Harris, K.~Jarrod Millman, St{\'{e}}fan~J. van~der Walt, Ralf
  Gommers, Pauli Virtanen, David Cournapeau, Eric Wieser, Julian Taylor,
  Sebastian Berg, Nathaniel~J. Smith, Robert Kern, Matti Picus, Stephan Hoyer,
  Marten~H. van Kerkwijk, Matthew Brett, Allan Haldane, Jaime~Fern{\'{a}}ndez
  del R{\'{i}}o, Mark Wiebe, Pearu Peterson, Pierre G{\'{e}}rard-Marchant,
  Kevin Sheppard, Tyler Reddy, Warren Weckesser, Hameer Abbasi, Christoph
  Gohlke, and Ravis~E. Oliphant.
\newblock {A}rray {P}rogramming with {NumPy}.
\newblock {\em Nature}, 585(7825):357--362, 2020.

\bibitem{honda2018deleterious}
Takayuki Honda, Hiroyuki Sakashita, Kyohei Masai, Hirohiko Totsuka, Noriko
  Motoi, Masashi Kobayashi, Takumi Akashi, Sachiyo Mimaki, Katsuya Tsuchihara,
  Suenori Chiku, et~al.
\newblock {D}eleterious {P}ulmonary {S}urfactant {S}ystem {G}ene {M}utations in
  {L}ung {A}denocarcinomas {A}ssociated with {U}sual {I}nterstitial
  {P}neumonia.
\newblock {\em {JCO Precision Oncology}}, 2:1--24, 2018.

\bibitem{hu2014trps1}
Jing Hu, Peng Su, Ming Jia, Xiaojuan Wu, Hui Zhang, Weiwei Li, and Gengyin
  Zhou.
\newblock {TRPS1} expression promotes angiogenesis and affects {VEGFA}
  expression in breast cancer.
\newblock {\em Experimental Biology and Medicine}, 239(4):423--429, 2014.

\bibitem{kukurba2015rna}
Kimberly~R Kukurba and Stephen~B Montgomery.
\newblock {RNA} {S}equencing and {A}nalysis.
\newblock {\em Cold Spring Harbor Protocols}, 2015(11):951--969, 2015.

\bibitem{le2018supervised}
Lei Le, Andrew Patterson, and Martha White.
\newblock {S}upervised autoencoders: {I}mproving generalization performance
  with unsupervised regularizers.
\newblock In {\em Advances in Neural Information Processing Systems}, pages
  107--117, 2018.

\bibitem{louizos2016variational}
Christos Louizos, Kevin Swersky, Yujia Li, Max Welling, and Richard~S Zemel.
\newblock The {V}ariational {F}air {A}utoencoder.
\newblock In {\em Proceedings of the International Conference on Learning
  Representations}, 2016.

\bibitem{makhzani2015adversarial}
Alireza Makhzani, Jonathon Shlens, Navdeep Jaitly, Ian Goodfellow, and Brendan
  Frey.
\newblock {A}dversarial {A}utoencoders.
\newblock {\em arXiv preprint arXiv:1511.05644}, 2015.

\bibitem{mcinnes2018umap}
Leland McInnes, John Healy, Nathaniel Saul, and Lukas Gro{\ss}berger.
\newblock {UMAP}: {U}niform {M}anifold {A}pproximation and {P}rojection for
  {D}imension {R}eduction.
\newblock {\em Journal of Open Source Software}, 3(29):861, 2018.

\bibitem{nathan2016germline}
Nadia Nathan, Violaine Giraud, Cl{\'e}ment Picard, Hilario Nunes, Florence
  Dastot-Le~Moal, Bruno Copin, Laurie Galeron, Alice De~Ligniville, Nathalie
  Kuziner, Martine Reynaud-Gaubert, et~al.
\newblock {G}ermline {SFTPA1} mutation in familial idiopathic interstitial
  pneumonia and lung cancer.
\newblock {\em Human Molecular Genetics}, 25(8):1457--1467, 2016.

\bibitem{pedregosa2011scikit}
Fabian Pedregosa, Ga{\"e}l Varoquaux, Alexandre Gramfort, Vincent Michel,
  Bertrand Thirion, Olivier Grisel, Mathieu Blondel, Peter Prettenhofer, Ron
  Weiss, Vincent Dubourg, et~al.
\newblock scikit-learn: {M}achine learning in {P}ython.
\newblock {\em The Journal of Machine Learning Research}, 12:2825--2830, 2011.

\bibitem{radvanyi2005gene}
Laszlo Radvanyi, Devender Singh-Sandhu, Scott Gallichan, Corey Lovitt, Artur
  Pedyczak, Gustavo Mallo, Kurt Gish, Kevin Kwok, Wedad Hanna, Judith Zubovits,
  et~al.
\newblock {T}he gene associated with trichorhinophalangeal syndrome in humans
  is overexpressed in breast cancer.
\newblock {\em Proceedings of the National Academy of Sciences},
  102(31):11005--11010, 2005.

\bibitem{ranzato2008semi}
Marc'Aurelio Ranzato and Martin Szummer.
\newblock {S}emi-supervised learning of compact document representations with
  deep networks.
\newblock In {\em Proceedings of the International Conference on Machine
  Learning}, pages 792--799, 2008.

\bibitem{shen2013epigenetic}
Hui Shen, Brooke~L. Fridley, Honglin Song, Kate Lawrenson, Julie~M. Cunningham,
  Susan~J. Ramus, Mine~S. Cicek, Jonathan Tyrer, Douglas Stram, Melissa~C.
  Larson, et~al.
\newblock {E}pigenetic analysis leads to identification of {HNF1B} as a
  subtype-specific susceptibility gene for ovarian cancer.
\newblock {\em Nature Communications}, 4(1):1--10, 2013.

\bibitem{sin2013pro}
Don~D. Sin, C.~Martin Tammemagi, Stephen Lam, Matt~J. Barnett, Xiaobo Duan,
  Anthony Tam, Heidi Auman, Ziding Feng, Gary~E. Goodman, Samir Hanash, et~al.
\newblock {P}ro-surfactant protein {B} as a biomarker for lung cancer
  prediction.
\newblock {\em Journal of Clinical Oncology}, 31(36):4536, 2013.

\bibitem{talbot2020relating}
Austin Talbot.
\newblock {\em {R}elating {T}raits to {E}lectrophysiology using {F}actor
  {M}odels}.
\newblock PhD thesis, Duke University, 2020.

\bibitem{talbot2020supervised}
Austin Talbot, David Dunson, Kafui Dzirasa, and David Carlson.
\newblock {S}upervised {A}utoencoders {L}earn {R}obust {J}oint {F}actor
  {M}odels of {N}eural {A}ctivity.
\newblock {\em arXiv preprint arXiv:2004.05209}, 2020.

\bibitem{van2008visualizing}
Laurens Van~der Maaten and Geoffrey Hinton.
\newblock {V}isualizing {D}ata using {t-SNE}.
\newblock {\em Journal of Machine Learning Research}, 9(11):2579--2605, 2008.

\bibitem{virtanen2020fundamental}
Pauli Virtanen, Ralf Gommers, Travis~E. Oliphant, Matt Haberland, Tyler Reddy,
  David Cournapeau, Evgeni Burovski, Pearu Peterson, Warren Weckesser, Jonathan
  Bright, St{\'e}fan~J. {van der Walt}, Matthew Brett, Joshua Wilson, K.~Jarrod
  Millman, Nikolay Mayorov, Andrew R.~J. Nelson, Eric Jones, Robert Kern, Eric
  Larson, C.~J. Carey, {\.I}lhan Polat, Yu~Feng, Eric~W. Moore, Jake
  {VanderPlas}, Denis Laxalde, Josef Perktold, Robert Cimrman, Ian Henriksen,
  E.~A. Quintero, Charles~R. Harris, Anne~M. Archibald, Ant{\^o}nio~H. Ribeiro,
  Fabian Pedregosa, Paul {van Mulbregt}, and {SciPy 1.0 Contributors}.
\newblock {SciPy} 1.0: {F}undamental {A}lgorithms for {S}cientific {C}omputing
  in {P}ython.
\newblock {\em Nature Methods}, 17:261--272, 2020.

\bibitem{wang2009genetic}
Yongyu Wang, Phillip~J. Kuan, Chao Xing, Jennifer~T. Cronkhite, Fernando
  Torres, Randall~L Rosenblatt, J.~Michael DiMaio, Lisa~N. Kinch, Nick~V.
  Grishin, and Christine~Kim Garcia.
\newblock {G}enetic {D}efects in {S}urfactant {P}rotein {A2} are {A}ssociated
  with {P}ulmonary {F}ibrosis and {L}ung {C}ancer.
\newblock {\em {The American Journal of Human Genetics}}, 84(1):52--59, 2009.

\bibitem{weinstein2013cancer}
John~N. Weinstein, Eric~A. Collisson, Gordon~B. Mills, Kenna R.~Mills Shaw,
  Brad~A. Ozenberger, Kyle Ellrott, Ilya Shmulevich, Chris Sander, and
  Joshua~M. Stuart.
\newblock {T}he {C}ancer {G}enome {A}tlas {P}an-{C}ancer {A}nalysis {P}roject.
\newblock {\em Nature Genetics}, 45(10):1113--1120, 2013.

\bibitem{witwicki2018trps1}
Robert~M. Witwicki, Muhammad~B. Ekram, Xintao Qiu, Michalina Janiszewska,
  Shaokun Shu, Mijung Kwon, Anne Trinh, Elizabeth Frias, Nadire Ramadan, Greg
  Hoffman, et~al.
\newblock {TRPS1} is a {L}ineage-{S}pecific {T}ranscriptional {D}ependency in
  {B}reast {C}ancer.
\newblock {\em Cell Reports}, 25(5):1255--1267, 2018.

\bibitem{wu2014central}
Lele Wu, Yuzhi Wang, Yan Liu, Shiyi Yu, Hao Xie, Xingjuan Shi, Sheng Qin, Fei
  Ma, Tuan~Zea Tan, Jean~Paul Thiery, et~al.
\newblock {A} central role for {TRPS1} in the control of cell cycle and cancer
  development.
\newblock {\em Oncotarget}, 5(17):7677, 2014.

\bibitem{xiao2017eight}
Jian Xiao, Xiaoxiao Lu, Xi~Chen, Yong Zou, Aibin Liu, Wei Li, Bixiu He, Shuya
  He, and Qiong Chen.
\newblock {E}ight potential biomarkers for distinguishing between lung
  adenocarcinoma and squamous cell carcinoma.
\newblock {\em Oncotarget}, 8(42):71759, 2017.

\bibitem{yu2006supervised}
Shipeng Yu, Kai Yu, Volker Tresp, Hans-Peter Kriegel, and Mingrui Wu.
\newblock Supervised {P}robabilistic {P}rincipal {C}omponent {A}nalysis.
\newblock In {\em Proceedings of the 12th ACM SIGKDD International Conference
  on Knowledge Discovery and Data Mining}, pages 464--473, 2006.

\bibitem{zhang2016augmenting}
Yuting Zhang, Kibok Lee, and Honglak Lee.
\newblock {A}ugmenting {S}upervised {N}eural {N}etworks with {U}nsupervised
  {O}bjectives for {L}arge-{S}cale {I}mage {C}lassification.
\newblock In {\em Proceedings of the International Conference on Machine
  Learning}, pages 612--621. PMLR, 2016.

\end{thebibliography}

\newpage

\appendix
\counterwithin{figure}{section}
\counterwithin{table}{section}

\section{AugmentedPCA Derivations and Analytic Solutions}
\label{sec:derivations}


Here, we derive the solutions for $\mathbf{W}$ and $\mathbf{D}$ in terms of matrix decompositions as well as the solution for $\mathbf{A}$ in the case of encoded approximate inference. Solutions for all approximate inference strategies are derived with respects to SAPCA objectives, and then subsequently shown that the solutions for AAPCA objectives are found similarly.

\subsection{Local Approximate Inference}

The objective of supervised AugmentedPCA (SAPCA) with local approximate inference is
\begin{equation}
\min_{\mathbf{W}, \mathbf{D}, \mathbf{S}} \norm{\mathbf{X} - \mathbf{W}\mathbf{S}}_F^2 + \mu\norm{\mathbf{Y} - \mathbf{D}\mathbf{S}}_F^2,
\label{eq:supp_lSAPCA_obj}
\end{equation}
where $\mathbf{W}$ and $\mathbf{D}$ are the loadings relating the factors to the primary and the loadings relating factors to the augmenting data, respectively. The above objective can be rewritten in terms of traces as
\begin{align}
L = \min_{\mathbf{W}, \mathbf{D}, \mathbf{S}} &\tr((\mathbf{X} - \mathbf{W}\mathbf{S})^{\intercal} (\mathbf{X} - \mathbf{W}\mathbf{S})) + \mu\tr((\mathbf{Y} - \mathbf{D}\mathbf{S})^{\intercal} (\mathbf{Y} - \mathbf{D}\mathbf{S})), \\
L = \min_{\mathbf{W}, \mathbf{D}, \mathbf{S}} &\tr(\mathbf{X}^{\intercal}\mathbf{X} - 2\mathbf{X}^{\intercal}\mathbf{W}\mathbf{S} + \mathbf{S}^{\intercal}\mathbf{W}^{\intercal}\mathbf{W}\mathbf{S}) \\ &- \mu\tr(\mathbf{Y}^{\intercal}\mathbf{Y} - 2\mathbf{Y}^{\intercal}\mathbf{D}\mathbf{S} + \mathbf{S}^{\intercal}\mathbf{D}^{\intercal}\mathbf{D}\mathbf{S}). \notag
\end{align}
We now compute the partial derivatives of the objective $L$ with respects to $\mathbf{W}$, $\mathbf{D}$, and $\mathbf{S}$. The partial derivative with respect to $\mathbf{W}$ is
\begin{align}
0 = \frac{\partial L}{\partial \mathbf{W}} &= -2\mathbf{X}\mathbf{S}^{\intercal} + 2\mathbf{W}\mathbf{S}\mathbf{S}^{\intercal}, \\
2\mathbf{X}\mathbf{S}^{\intercal} &= 2\mathbf{W}\mathbf{S}\mathbf{S}^{\intercal}, \\
\mathbf{W} &= \mathbf{X}\mathbf{S}^{\intercal}(\mathbf{S}\mathbf{S}T)^{-1}.
\end{align}
The partial derivative of the objective $L$ with respect to $\mathbf{D}$ is
\begin{align}
0 = \frac{\partial L}{\partial \mathbf{D}} &= -2\mu\mathbf{Y}\mathbf{S}^{\intercal} + 2\mu\mathbf{D}\mathbf{S}\mathbf{S}^{\intercal},\\
2\mathbf{Y}\mathbf{S}^{\intercal} &= 2\mathbf{D}\mathbf{S}\mathbf{S}^{\intercal},\\
\mathbf{D} &= \mathbf{Y}\mathbf{S}^{\intercal}(\mathbf{S}\mathbf{S}^{\intercal})^{-1}.
\end{align}
The partial derivative of the objective $L$ with respect to $\mathbf{S}$ is
\begin{align}
0 = \frac{\partial L}{\partial \mathbf{S}} &= -2\mathbf{W}^{\intercal}\mathbf{X} + 2\mathbf{W}^{\intercal}\mathbf{W}\mathbf{S} - 2\mu\mathbf{D}^{\intercal}\mathbf{Y} + 2\mu\mathbf{D}^{\intercal}\mathbf{D}\mathbf{S}, \\
\mathbf{W}^{\intercal}\mathbf{X} + \mu\mathbf{D}^{\intercal}\mathbf{Y} &= \mathbf{W}^{\intercal}\mathbf{W}\mathbf{S} + \mu\mathbf{D}^{\intercal}\mathbf{D}\mathbf{S}, \\
\mathbf{S} &= (\mathbf{W}^{\intercal}\mathbf{W} + \mu\mathbf{D}^{\intercal}\mathbf{D})^{-1} (\mathbf{W}^{\intercal}\mathbf{X} + \mu\mathbf{D}^{\intercal}\mathbf{Y}).
\end{align}
We first solve for a single component. Therefore, primary objective loadings, augmenting objective loadings, and components are represented as vectors $\mathbf{w}$, $\mathbf{d}$, and $\mathbf{s}$, respectively. For a single component, Equation \ref{eq:supp_lSAPCA_obj} can be rewritten as
\begin{align}
\mathbf{s} &= (\norm{\mathbf{w}}^2 + \mu\norm{\mathbf{d}}^2)^{-1} (\mathbf{w}^{\intercal}\mathbf{X} + \mu\mathbf{d}^{\intercal}\mathbf{Y})
\end{align}
Let $\alpha=(\norm{\mathbf{w}}^2 + \mu\norm{\mathbf{d}}^d)^{-1}$ and
$\gamma=(\mathbf{s}\mathbf{s}^{\intercal})^{-1}$. Since $\mathbf{s}$ is a vector both $\alpha$ and $\gamma$ are constant scalars, we can rewrite our fixed point equations as
\begin{align}
\mathbf{w} &= \gamma \mathbf{X}\mathbf{s}^{\intercal}, \\
\mathbf{d} &= \gamma \mathbf{Y}\mathbf{s}^{\intercal}, \\
\mathbf{s} &= \alpha (\mathbf{w}^{\intercal}\mathbf{X} + \mu\mathbf{d}^{\intercal}\mathbf{Y}).
\end{align}
When we substitute $\mathbf{s}$ into the above equations for $\mathbf{w}$ and $\mathbf{d}$ we end up with
\begin{align}
\mathbf{w} &= \alpha\gamma(\mathbf{X}\mathbf{X}^{\intercal}\mathbf{w} + \mu\mathbf{X}\mathbf{Y}^{\intercal}\mathbf{d}), \\
\mathbf{d} &= \alpha\gamma(\mathbf{Y}\mathbf{X}^{\intercal}\mathbf{w} + \mu\mathbf{Y}\mathbf{Y}^{\intercal}\mathbf{d}).
\end{align}
Because $\alpha$ and $\gamma$ are scalars, the solution for $[\mathbf{W}^{\intercal}, \mathbf{D}^{\intercal}]^{\intercal}$  correspond to the eigenvectors associated with the largest eigenvalues of the matrix
\[
\mathbf{B}_{\text{SL}} = \begin{bmatrix}
\mathbf{X}\mathbf{X}^{\intercal} & \mu \mathbf{X}\mathbf{Y}^{\intercal} \\
\mathbf{Y}\mathbf{X}^{\intercal} & \mu \mathbf{Y}\mathbf{Y}^{\intercal}
\end{bmatrix}.
\]
We can find the solutions additional factors by iteratively by finding a new vector orthogonal to the previous factor solutions.

Additionally, the only difference between the objective equation for adversarial AugmentedPCA (AAPCA) with local approximate inference and the objective equation for SAPCA with local approximate inference is a sign change on the augmenting objective. Thus, the solutions for $[\mathbf{W}^{\intercal}, \mathbf{D}^{\intercal}]^{\intercal}$ in the case of AAPCA with local approximate inference correspond to the eigenvectors associated with the largest eigenvalues of the matrix
\[
\mathbf{B}_{\text{SE}} = \begin{bmatrix}
\mathbf{X}\mathbf{X}^{\intercal} & -\mu \mathbf{X}\mathbf{Y}^{\intercal} \\
\mathbf{Y}\mathbf{X}^{\intercal} & -\mu \mathbf{Y}\mathbf{Y}^{\intercal}
\end{bmatrix}.
\]

\subsection{Encoded Approximate Inference}

The objective of SAPCA with encoded approximate inference is 
\begin{equation}
\min_{\mathbf{W}, \mathbf{A}}\max_{\mathbf{D}} \norm{\mathbf{X} - \mathbf{W}\mathbf{A}\mathbf{X}}_F^2 + \mu\norm{\mathbf{Y} - \mathbf{D}\mathbf{A}\mathbf{X}}_F^2,
\label{eq:supp_eSAPCA_obj}
\end{equation}
where $\mathbf{A}$ is a linear mapping, and $\mathbf{W}$ and $\mathbf{D}$ retain their meanings as the loadings relating the factors to the primary and the loadings relating factors to the augmenting data, respectively. The above objective can be rewritten in terms of traces as
\begin{align}
L = \min_{\mathbf{W}, \mathbf{A}}\max_{\mathbf{D}} &\tr((\mathbf{X} - \mathbf{W}\mathbf{A}\mathbf{X})^{\intercal}(\mathbf{X} - \mathbf{W}\mathbf{A}\mathbf{X})) + \mu\tr((\mathbf{Y} - \mathbf{D}\mathbf{A}\mathbf{X})^{\intercal}(\mathbf{Y} - \mathbf{D}\mathbf{A}\mathbf{X})),\\
L = \min_{\mathbf{W}, \mathbf{A}}\max_{\mathbf{D}} &\tr(\mathbf{X}^{\intercal}\mathbf{X} - 2\mathbf{X}^{\intercal}\mathbf{W}\mathbf{A}\mathbf{X} + \mathbf{X}^{\intercal}\mathbf{A}^{\intercal}\mathbf{W}^{\intercal}\mathbf{W}\mathbf{A}\mathbf{X}) \\ &+ \mu\tr(\mathbf{Y}^{\intercal}\mathbf{Y} - 2\mathbf{Y}^{\intercal}\mathbf{D}\mathbf{A}\mathbf{X} + \mathbf{X}^{\intercal}\mathbf{A}^{\intercal}\mathbf{D}^{\intercal}\mathbf{D}\mathbf{A}\mathbf{X}). \notag
\end{align}
We now compute the partial derivatives of the objective $L$ with respects to $\mathbf{W}$, $\mathbf{S}$ and $\mathbf{A}$. The partial derivative with respect to $\mathbf{W}$ is
\begin{align}
0 = \frac{\partial L}{\partial \mathbf{W}} &= -2\mathbf{X}\mathbf{X}^{\intercal}\mathbf{A}^{\intercal} + 2\mathbf{W}\mathbf{A}\mathbf{X}\mathbf{X}^{\intercal}\mathbf{A}^{\intercal}, \\
2\mathbf{X}\mathbf{X}^{\intercal}\mathbf{A}^{\intercal} &= 2\mathbf{W}\mathbf{A}\mathbf{X}\mathbf{X}^{\intercal}\mathbf{A}^{\intercal}, \\
\mathbf{W} &= \mathbf{X}\mathbf{X}^{\intercal}\mathbf{A}^{\intercal}(\mathbf{A}\mathbf{X}\mathbf{X}^{\intercal}\mathbf{A}^{\intercal})^{-1}.
\end{align}
The partial derivative with respect to $\mathbf{D}$ is
\begin{align}
0 = \frac{\partial L}{\partial \mathbf{D}} &= -2\mu\mathbf{Y}\mathbf{X}^{\intercal}\mathbf{A}^{\intercal} + 2\mu\mathbf{D}\mathbf{A}\mathbf{X}\mathbf{X}^{\intercal}\mathbf{A}^{\intercal}, \\
2\mathbf{Y}\mathbf{X}^{\intercal}\mathbf{A}^{\intercal} &= 2\mathbf{D}\mathbf{A}\mathbf{X}\mathbf{X}^{\intercal}\mathbf{A}^{\intercal}, \\
\mathbf{D} &= \mathbf{Y}\mathbf{X}^{\intercal}\mathbf{A}^{\intercal}(\mathbf{A}\mathbf{X}\mathbf{X}^{\intercal}\mathbf{A}^{\intercal})^{-1}.
\end{align}
The partial derivative with respect to $\mathbf{A}$ is
\begin{align}
0 = \frac{\partial L}{\partial \mathbf{A}} = &-2\mathbf{W}^{\intercal}\mathbf{X}\mathbf{X}^{\intercal} + 2\mathbf{W}^{\intercal}\mathbf{W}\mathbf{A}\mathbf{X}\mathbf{X}^{\intercal} \\
&- 2\mu\mathbf{D}^{\intercal}\mathbf{Y}\mathbf{X}^{\intercal} + 2\mu\mathbf{D}^{\intercal}\mathbf{D}\mathbf{A}\mathbf{X}\mathbf{X}^{\intercal}, \notag \\
2\mathbf{W}^{\intercal}\mathbf{X}\mathbf{X}^{\intercal} + 2\mu\mathbf{D}^{\intercal}\mathbf{Y}\mathbf{X}^{\intercal} = &2\mathbf{W}^{\intercal}\mathbf{W}\mathbf{A}\mathbf{X}\mathbf{X}^{\intercal} + 2\mu\mathbf{D}^{\intercal}\mathbf{D}\mathbf{A}\mathbf{X}\mathbf{X}^{\intercal}, \\
(\mathbf{W}^{\intercal}\mathbf{X} + \mu\mathbf{D}^{\intercal}\mathbf{Y})\mathbf{X}^{\intercal} = &(\mathbf{W}^{\intercal}\mathbf{W} + \mu\mathbf{D}^{\intercal}\mathbf{D})\mathbf{A}\mathbf{X}\mathbf{X}^{\intercal}, \\
\mathbf{A} = &(\mathbf{W}^{\intercal}\mathbf{W} + \mu\mathbf{D}^{\intercal}\mathbf{D})^{-1} (\mathbf{W}^{\intercal}\mathbf{X} + \mu\mathbf{D}^{\intercal}\mathbf{Y}) \mathbf{X}^{\intercal}(\mathbf{X}\mathbf{X}^{\intercal})^{-1}.
\label{eq:big_a_eq}
\end{align}
We now solve for a single component. Primary objective loadings, augmenting objective loadings, and linear encoding mapping are represented as vectors $\mathbf{w}$, $\mathbf{d}$, and $\mathbf{a}$, respectively. For a single component, Equation \ref{eq:big_a_eq} can be rewritten as
\begin{align}
\mathbf{a} &= (\norm{\mathbf{w}}^2 + \mu\norm{\mathbf{d}}^2)^{-1} (\mathbf{w}^{\intercal}\mathbf{X} + \mu\mathbf{d}^{\intercal}\mathbf{Y}) \mathbf{X}^{\intercal}(\mathbf{X}\mathbf{X}^{\intercal})^{-1}.
\end{align}
Once again, we let $\alpha=(\norm{\mathbf{w}}^2 + \mu\norm{\mathbf{d}}^2)^{-1}$ and $\gamma=(\mathbf{s}\mathbf{s}^{\intercal})^{-1}$. We can rewrite our fixed point equations as
\begin{align}
\mathbf{w} &= \gamma\mathbf{X}\mathbf{X}^{\intercal}\mathbf{a}^{\intercal}, \\
\mathbf{d} &= \gamma\mathbf{Y}\mathbf{X}^{\intercal}\mathbf{a}^{\intercal}, \\
\mathbf{a} &= \alpha(\mathbf{w}^{\intercal}\mathbf{X} + \mu\mathbf{d}^{\intercal}\mathbf{Y})\mathbf{X}^{\intercal}(\mathbf{X}\mathbf{X}^{\intercal})^{-1}.
\end{align}
When we substitute $\mathbf{a}$ into the above equations for $\mathbf{w}$ and $\mathbf{d}$ we end up with
\begin{align}
\mathbf{w} &= \alpha\gamma\mathbf{X}\mathbf{X}^{\intercal}(\mathbf{X}\mathbf{X}^{\intercal})^{-1} \mathbf{X}(\mathbf{X}^{\intercal}\mathbf{w} + \mu\mathbf{Y}^{\intercal}\mathbf{d}), \\
\mathbf{d} &= \alpha\gamma\mathbf{Y}\mathbf{X}^{\intercal}(\mathbf{X}\mathbf{X}^{\intercal})^{-1} \mathbf{X}(\mathbf{X}^{\intercal}\mathbf{w} + \mu\mathbf{Y}^{\intercal}\mathbf{d}).
\end{align}
Thus, the solution for $[\mathbf{W}^{\intercal}, \mathbf{D}^{\intercal}]^{\intercal}$  can be found as the eigenvectors associated with the largest eigenvalues of the matrix correspond to the eigenvectors associated with the largest eigenvalues of the matrix
\[
\mathbf{B}_{\text{AL}} = \begin{bmatrix}
\mathbf{X}\mathbf{X}^{\intercal} & \mu \mathbf{X}\mathbf{Y}^{\intercal} \\
\mathbf{Y}\mathbf{X}^{\intercal} & \mu \mathbf{Y}\textbf{P}_{\mathbf{X}}\mathbf{Y}^{\intercal}
\end{bmatrix},
\]
where $\mathbf{P}_{\mathbf{X}} = \mathbf{X}(\mathbf{X}\mathbf{X}^{\intercal})^{-1}\mathbf{X}^{\intercal}$ is the projection matrix on $\mathbf{X}$. We can find the solutions additional factors by iteratively by finding a new vector orthogonal to the previous factor solutions.

As noted in the local approximate inference derivation, the only difference between the objective equation for adversarial AugmentedPCA (AAPCA) with encoded approximate inference and the objective equation for SAPCA with encoded approximate inference is a sign change on the augmenting objective. Thus, the solutions for $[\mathbf{W}^{\intercal}, \mathbf{D}^{\intercal}]^{\intercal}$ in the case of AAPCA with encoded approximate inference correspond to the eigenvectors associated with the largest eigenvalues of the matrix
\[
\mathbf{B}_{\text{AE}} = \begin{bmatrix}
\mathbf{X}\mathbf{X}^{\intercal} & -\mu \mathbf{X}\mathbf{Y}^{\intercal} \\
\mathbf{Y}\mathbf{X}^{\intercal} & -\mu \mathbf{Y}\mathbf{P}_{\mathbf{X}}\mathbf{Y}^{\intercal}
\end{bmatrix}.
\]

\section{Randomized AugmentedPCA - Further Details}
\label{sec:rapca_add_details}

The work by Halko, Martinsson, and Tropp \cite{halko2011finding} discussed algorithms that use random sampling to compute approximate matrix decompositions. We leverage their proposed randomized algorithm for computing an approximate eigendecomposition, and in doing so trade exact analytic solutions for a speed up in computation. The steps for computing an approximate eigendecomposition are detailed in Algorithm \ref{alg:random_eig}. This randomized technique greatly decreases algorithmic complexity from $\mathcal{O}(P^3)$ in the analytic case to $\mathcal{O}(P^2 (k + s))$.
Here, $P$ represents the combined number of primary and augmenting features $p + q$, $k$ represents the chosen number of components, and $s$ is a parameter called the oversampling term that increases the fidelity of the approximation as it increases.

In brief, $\boldsymbol{\Omega} \in \mathbb{R}^{(p + q) \times (k + s)}$ is a random test matrix drawn from a standard Gaussian distribution, and $t$ represents the number of subspace iterations.  For test matrices ($\boldsymbol{\Omega}$) drawn from a standard Gaussian, the oversampling parameter is recommended to be a small constant, such as $s$ = 5 or $s$ = 10 \cite{halko2011finding}. Additionally, a power iteration scheme like the one depicted in the $\texttt{for}$ loop in Algorithm \ref{alg:random_eig} helps reduce the approximation gap of the randomized matrix decomposition. For AugmentedPCA model implementations, we set both the default values of $s$ and $t$ to 5. For further details on randomized algorithms for approximate matrix decompositions, we refer to the work by Halko, Martinsson, and Tropp \cite{halko2011finding}.

\begin{algorithm}[H]
    \KwData{Construct decomposition matrix $\mathbf{B} \in \mathbb{R}^{(p + q) \times (p + q)}$ from primary and augmenting data matrices $\mathbf{X} \in \mathbb{R}^{n \times p}$ and $\mathbf{Y} \in \mathbb{R}^{n \times q}$ according to Equation \ref{eq:sapca_loc_dec_mat}, \ref{eq:sapca_enc_dec_mat}, \ref{eq:aapca_loc_dec_mat}, or \ref{eq:aapca_enc_dec_mat}, depending on model type and chosen inference strategy.}
    \KwRequire{Number of components $k > 0$, oversampling parameter $s \geq 0$, number of power iterations parameter $t \geq 0$, $k + s < \min(n, p)$} \vspace{2mm}
    $\boldsymbol{\Omega} \in \mathbb{R}^{(p + q) \times (k + s)} \sim \mathcal{N}(0, 1)$ \\
    $\mathbf{G} = \mathbf{B} \boldsymbol{\Omega}$ \\
    Compute orthogonal column basis $\mathbf{Q}$ via QR factorization: $\mathbf{Q}, \mathbf{R} = \mathbf{G}$ \\ \vspace{2mm}
    \For{$i \leftarrow 1:1:t$}{
        $\widetilde{\mathbf{G}} = \mathbf{B}^{\intercal} \mathbf{Q}$ \\
        Compute orthogonal column basis $\widetilde{\mathbf{Q}}$ via QR factorization: $\widetilde{\mathbf{Q}}, \widetilde{\mathbf{R}} = \widetilde{\mathbf{G}}$ \\
        $\mathbf{G} = \mathbf{B} \widetilde{\mathbf{Q}}$ \\
        Compute orthogonal column basis $\mathbf{Q}$ via QR factorization: $\mathbf{Q}, \mathbf{R} = \mathbf{G}$
    } \vspace{2mm}
    $\mathbf{B}_{\text{approx}} = \mathbf{Q}^{\intercal} \mathbf{B} \mathbf{Q}$ \\
    Compute eigendecomposition $\mathbf{B}_{\text{approx}} = \mathbf{V} \mathbf{\Lambda} \mathbf{V}^{\intercal}$ \\
    $\mathbf{U} = \mathbf{Q} \mathbf{V}$ \\ \vspace{2mm}
    \Return $\mathbf{U}$ \\ \vspace{2mm}
    \caption{Approximate AugmentedPCA eigendecomposition algorithm.}
    \label{alg:random_eig}
\end{algorithm}

\section{AugmentedPCA Loadings Interpretation - Additional Results}
\label{sec:loadings_interpretation_additional}

Here, we learn SAPCA representations specific to the different types of cancer in the RNA-seq dataset by enforcing components to be predictive of binary cancer labels for each of the five cancer types. Here, the augmenting supervised data corresponds to a matrix $\mathbf{Y}^{\intercal} \in \mathbb{R}^{801 \times 1}$ for a given cancer type. After fitting SAPCA models, elements of the loading vector corresponding to the first principal component were sorted according to descending magnitudes to identify the top genes that were implicated in each specific type of cancer.

\begin{table}[t!]
    \centering
    \caption{Top 10 genes associated with COAD according to learned SAPCA representations. Genes were sorted according to the largest magnitude elements of SAPCA component 1 loading.}
    \smallskip
    \begin{tabular}{|c|c|c|}
    \hline
    \textbf{Gene Symbol} & \textbf{Location}  & \textbf{Loading Value} \\
    \hline
    ANKRD40CL & chr17: 50.76-50.77 Mb & 0.0261 \\
    CDX1 & chr5: 150.17-150.18 Mb & 0.0256 \\
    MEP1A & chr6: 46.79-46.85 Mb & 0.0245 \\
    GUCY2C & chr12: 14.61-14.70 Mb & 0.0244 \\
    MYO1A & chr12: 57.03-57.05 Mb & 0.0244 \\
    FLJ32063 & chr2: 200.32-200.34 Mb & 0.0243 \\
    GPA33 & chr1: 167.05-167.17 Mb & 0.0242 \\
    NOX1 & chrX: 100.84-100.87 Mb & 0.0239 \\
    PDX1 & chr13: 27.92-27.93 Mb & 0.0238 \\
    RETNLB & chr3: 108.74-108.76 Mb & 0.0237 \\
    \hline
    \end{tabular}
    \label{tab:top_genes_coad}
\end{table}

\begin{table}[ht!]
    \centering
    \caption{Top 10 genes associated with KIRC according to learned SAPCA representations. Genes were sorted according to the largest magnitude elements of SAPCA component 1 loading.}
    \smallskip
    \begin{tabular}{|c|c|c|}
    \hline
    \textbf{Gene Symbol} & \textbf{Location}  & \textbf{Loading Value} \\
    \hline
    ACSM2A & chr16: 20.45-20.49 Mb & 0.0223 \\
    PRODH2 & chr19: 35.80-35.81 Mb & 0.0223 \\
    ACSM2B & chr16: 20.54-20.58 Mb & 0.0223 \\
    OVOL2 & chr20: 17.96-18.06 Mb & -0.0220 \\
    AGXT2 & chr5: 35.00-35.05 Mb & 0.0220 \\
    POU3F3 & chr2: 104.85-104.86 Mb & 0.0219 \\
    BHMT & chr5: 79.11-79.13 Mb & 0.0219 \\
    FXYD2 & chr11: 117.80-117.83 Mb & 0.0218 \\
    CDH16 & chr16: 66.91-66.92 Mb & 0.0218 \\
    SLC22A2 & chr6: 160.17-160.28 Mb & 0.0218 \\
    \hline
    \end{tabular}
    \label{tab:top_genes_kirc}
\end{table}

\begin{table}[t!]
    \centering
    \caption{Top 10 genes associated with PRAD according to learned SAPCA representations. Genes were sorted according to the largest magnitude elements of SAPCA component 1 loading.}
    \smallskip
    \begin{tabular}{|c|c|c|}
    \hline
    \textbf{Gene Symbol} & \textbf{Location}  & \textbf{Loading Value} \\
    \hline
    KLK2 & chr19: 50.86-50.88 Mb & 0.0243 \\
    KLK3 & chr19: 50.85-50.86 Mb & 0.0241 \\
    TMEFF2 & chr2: 191.95-192.19 & 0.0240 \\
    CHRNA2 & chr8: 27.46-27.48 & 0.0237 \\
    LMAN1L & chr15: 74.81-74.83 Mb & 0.0232 \\
    PCA3 & chr9: 76.69-76.86 Mb & 0.0231 \\
    PABPC1L2B & chrX: 73.003-73,006 Mb & 0.0230 \\
    PAGE4 & chrX: 49.83-49.83 Mb & 0.0230 \\
    SLC45A3 & chr1: 205.66-205.68 Mb & 0.0229 \\
    ACP3 & chr3: 132.32-132.37 Mb & 0.0229 \\
    \hline
    \end{tabular}
    \label{tab:top_genes_prad}
\end{table}

\begin{figure*}[t!]
  \centering
  \begin{subfigure}[]{0.325\textwidth}
    \centering
    \includegraphics[width=0.98\textwidth]{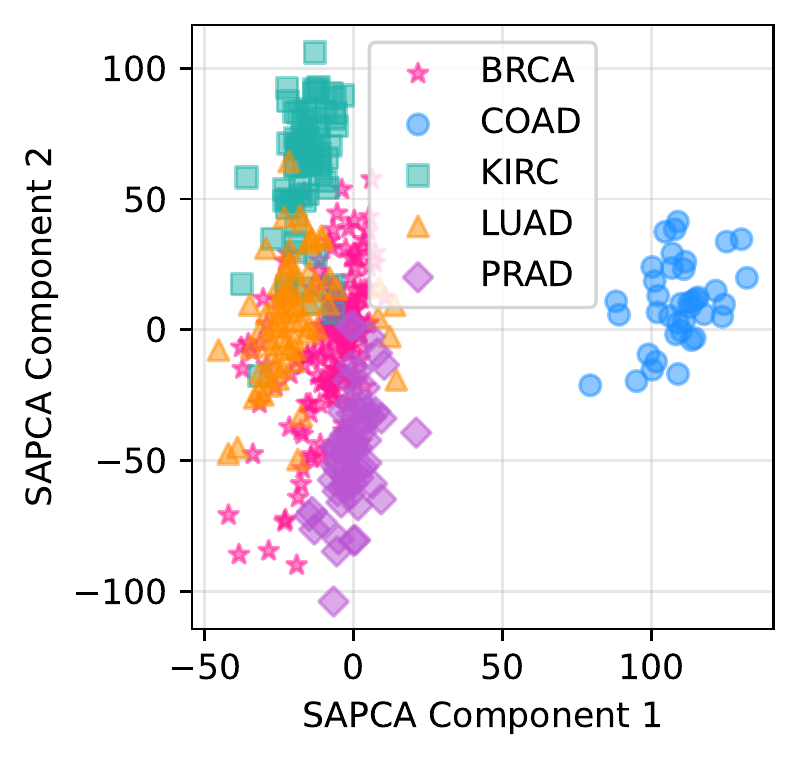}
    \caption{}
    \label{fig:SAPCA_cluster_coad}
  \end{subfigure}
  \begin{subfigure}[]{0.307\textwidth}
    \centering
    \includegraphics[width=0.98\textwidth]{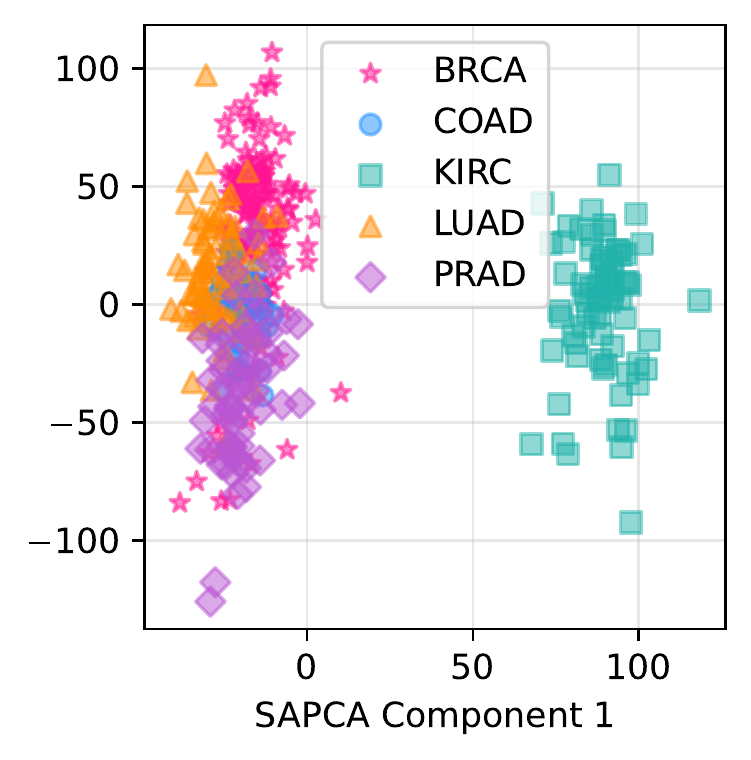}
    \caption{}
    \label{fig:SAPCA_cluster_kirc}
  \end{subfigure}
  \begin{subfigure}[]{0.307\textwidth}
    \centering
    \includegraphics[width=0.98\textwidth]{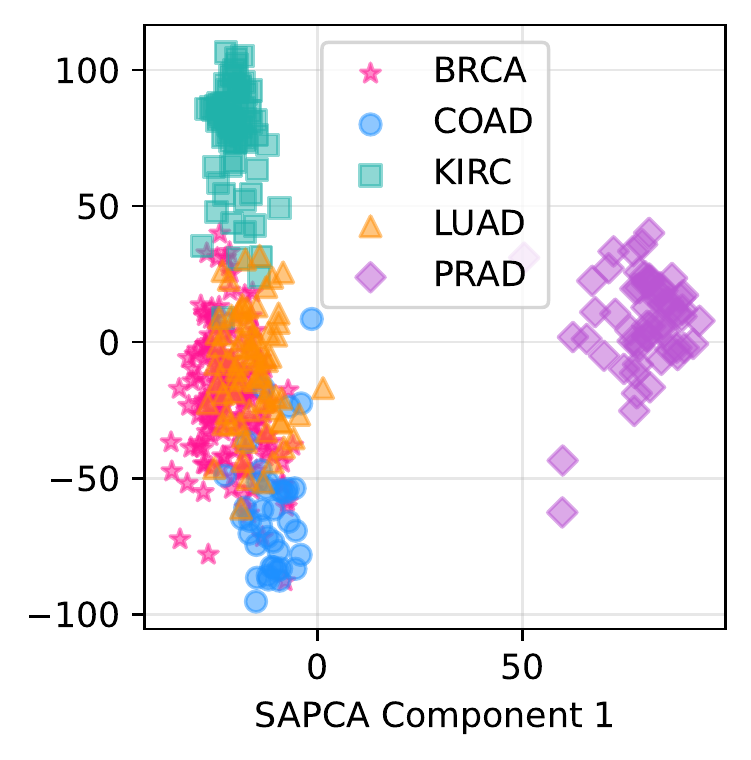}
    \caption{}
    \label{fig:SAPCA_cluster_prad}
  \end{subfigure}
  \caption{(a) Clustering of test set samples with SAPCA components enforced to be predictive of COAD labels. Supervision strength $\mu$ = $4 \cdot 10^{4}$. (b) Clustering of test set samples with SAPCA components enforced to be predictive of KIRC labels. Supervision strength $\mu$ = $4 \cdot 10^{4}$. (c) Clustering of test set samples with SAPCA components enforced to be predictive of PRAD labels. Supervision strength $\mu$ = $4 \cdot 10^{4}$.}
  \label{fig:SAPCA_loadings_cluster_additional}
\end{figure*}

In Tables \ref{tab:top_genes_coad}, \ref{tab:top_genes_kirc}, and \ref{tab:top_genes_prad}, we list the genes corresponding to the highest element magnitudes of SAPCA component 1 loading when enforcing components to be predictive of these cancers. Figure \ref{fig:SAPCA_loadings_cluster_additional} shows test set samples clustered in 2D SAPCA space when enforcing SAPCA components to be predictive of individual cancer type labels COAD, KIRC, and PRAD, respectively. We note that there is clear separation between COAD, KIRC, and PRAD samples and all other cancers as shown in the plots in Figure \ref{fig:SAPCA_cluster_coad}, \ref{fig:SAPCA_cluster_kirc}, and \ref{fig:SAPCA_cluster_prad}, respectively.

\section{Adversarial AugmentedPCA Example - Removal of Nuisance Variables in Image Data}
\label{sec:adv_apca_example}

\begin{figure*}[ht!]
  \centering
  \begin{subfigure}[]{0.99\textwidth}
    \centering
    \includegraphics[width=1.0\textwidth]{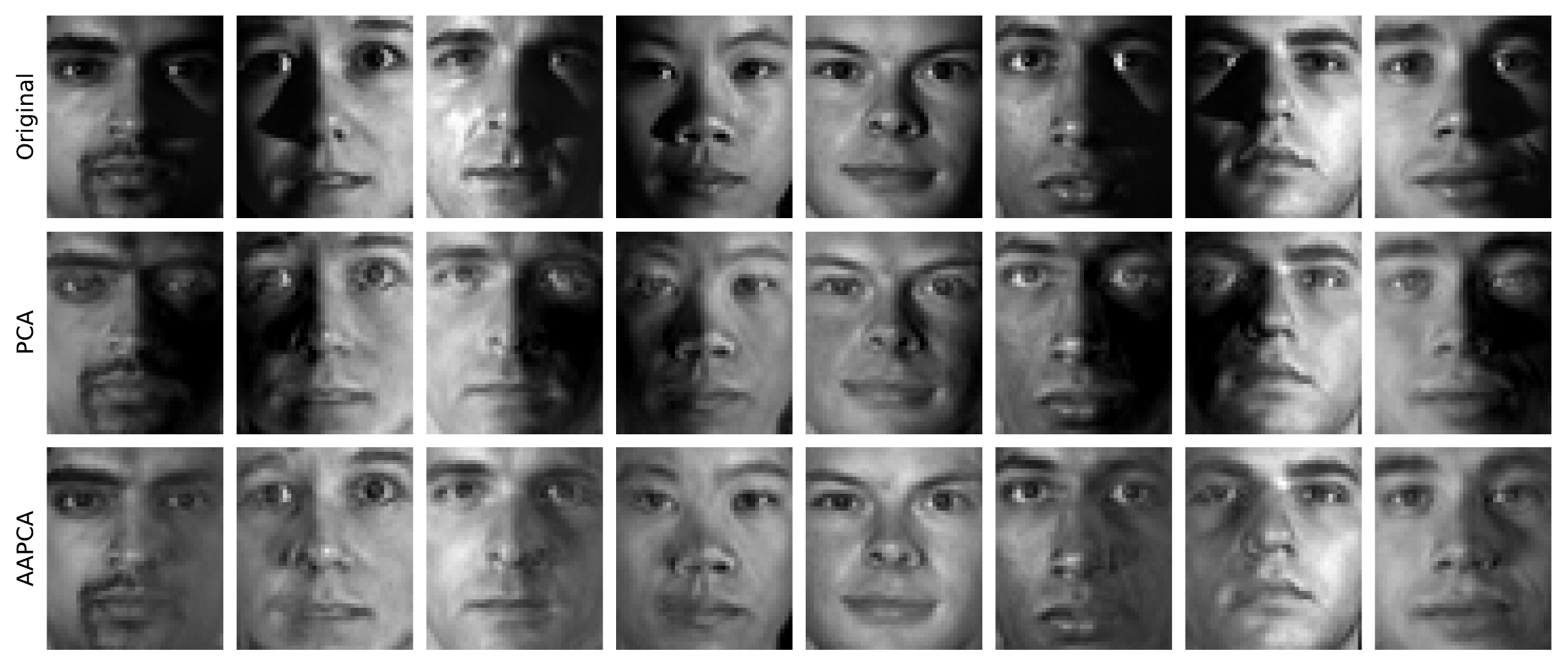}
    \caption{}
    \label{fig:face_recon}
  \end{subfigure}
  \begin{subfigure}[]{0.58\textwidth}
    \centering
    \includegraphics[width=1.0\textwidth]{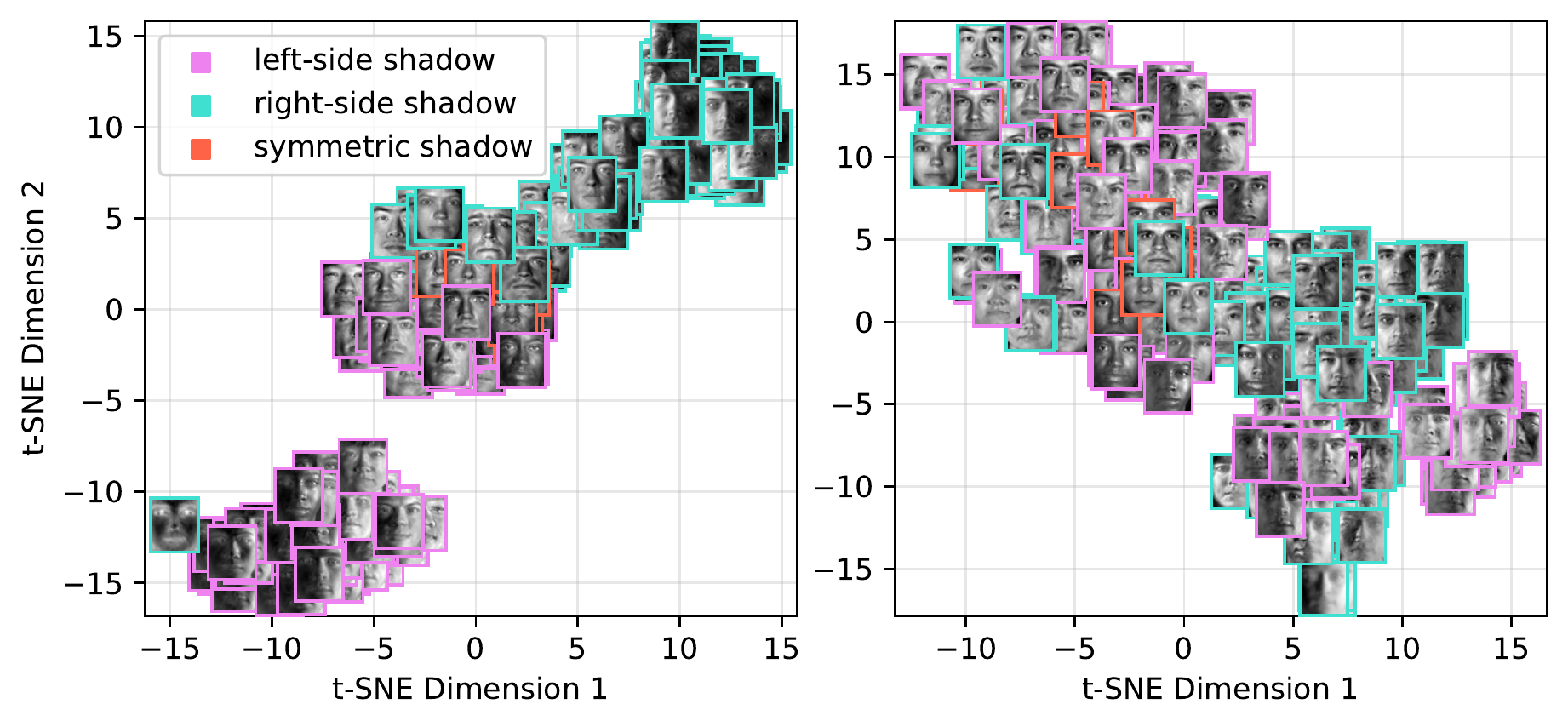}
    \caption{}
    \label{fig:tsne_cluster}
  \end{subfigure}
  \begin{subfigure}[]{0.37\textwidth}
    \centering
    \includegraphics[width=1.0\textwidth]{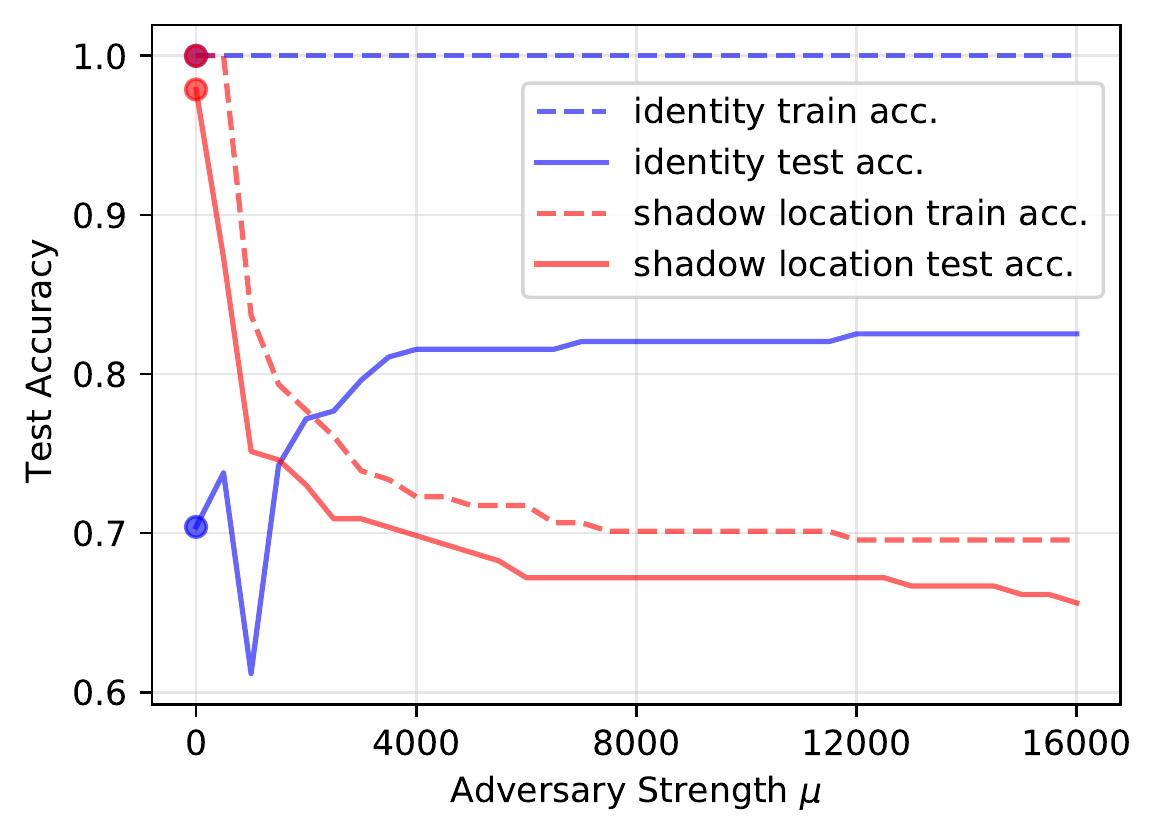}
    \caption{}
    \label{fig:face_classification}
  \end{subfigure}
  \label{fig:SAPCA_example}
  \caption{(a) Selected test set AAPCA reconstructions of shadow-invariant representations with adversarial strength $\mu$ = 20,000 (bottom row), PCA image reconstructions (middle row), and original images (top row). (b) t-SNE clustering of PCA image reconstructions (left) and AAPCA image reconstructions with adversarial strength $\mu$ = 20,000 (right). (c) Identity and shadow location classification as a function of adversary strength.}
\end{figure*}

We demonstrate the ability of AAPCA to create components invariant to concomitant data using images from the \href{http://vision.ucsd.edu/~leekc/ExtYaleDatabase/ExtYaleB.html}{Extended Yale Face Database B} \cite{georghiades2001from}. This dataset contains facial images of 38 human subjects taken with the light source at varying angles of azimuth and elevation, resulting in shadows cast across subject faces. Here, the nuisance variable is the variable lighting angles resulting in shadows that obscure parts of the image and by extension features of subject identity. We select a subset of 411 images in which only the azimuth of the light source was varied and elevation remained at a neutral 0 degrees above horizontal. We then use AAPCA with local approximate inference to create shadow-invariant representations.

First, a subset of 411 images was selected in which only the azimuth of the light source was varied and light source elevation remained at a neutral 0 degrees above horizontal. Images were down-sampled to be 0.25 times their original size, resulting in $42 \times 48$ images for a total of 2016 features per image. For each image, pixel intensities were scaled to be between 0 and 1. Images were flattened into row vectors of features and concatenated to form a matrix $\mathbf{X}^{\intercal} \in \mathbb{R}^{411 \times 2016}$ of primary data. Light source azimuth angle was considered the concomitant data, resulting in a concomitant data matrix $\mathbf{Y}^{\intercal} \in \mathbb{R}^{411 \times 1}$. Images were divided into a roughly 50-50 train-test split using the \texttt{train\_test\_split()} function from the scikit-learn package. We use AAPCA with 100 components to create shadow-invariant representations. Local inference is used as the approximate inference strategy since it is reasonable to assume access to concomitant data at test time. A logistic regression classifier was fit using the components derived from the training data and tested on the components derived from the test data for adversarial strengths $\mu$ = 0, 500, ..., 16,000.

Figure \ref{fig:face_recon} shows selected test set examples of AAPCA reconstructions (bottom row) compared to original images (top row). AAPCA reconstructions display noticeable shadow removal, thus demonstrating AAPCA's ability to produce nuisance-invariant representations. Figure \ref{fig:tsne_cluster} compares t-SNE clustering of PCA reconstructions of images and AAPCA reconstructions of images. PCA-reconstructions are grouped almost exclusively according to shadow location (left-side or right-side) in 2D space, while AAPCA-reconstructed images are grouped in a more shadow-invariant manner. Figure \ref{fig:face_classification} shows classification accuracy using AAPCA components to classify identity and shadow location as a function of adversary strength $\mu$. As the adversarial strength is increased, both training and test set classification accuracy of the nuisance variable (shadow location) decreases. For all adversary strengths, training set identity classification is 100\%. Initially, training on PCA representations results in a test set identity classification accuracy of 70\%. As adversary strength is increased, test set identity classification accuracy increases to 82\%, thus demonstrating the ability of AAPCA to mitigate the effects of domain shift due to concomitant influence.

\section{AugmentedPCA Logo}
\label{sec:apca_logo}

\begin{figure*}[ht!]
  \centering
  \includegraphics[width=1.0\textwidth]{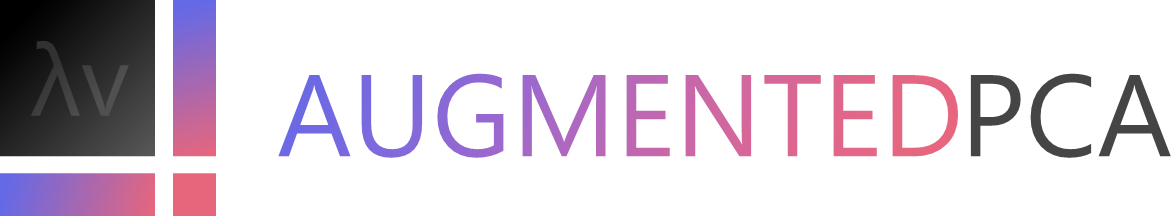}
  \caption{AugmentedPCA logo.}
  \label{fig:apca_logo}
\end{figure*}

\end{document}